\documentclass[sigconf, nonacm]{acmart}

\usepackage{booktabs}
\usepackage{graphicx}
\usepackage{subcaption}
\usepackage{amsmath,amssymb}
\usepackage{multirow}
\usepackage{xcolor}

\renewcommand\footnotetextcopyrightpermission[1]{}
\begin{document}

\title{HPSv3++: Scaling Reward Models Across the Full Spectrum of Diffusion Model Capabilities}

\author{Yijun Liu}
\affiliation{\institution{Tsinghua University}\country{ }}

\author{Jie Huang}
\affiliation{\institution{JD Explore Academy}\country{ }}

\author{Zeyue Xue}
\affiliation{\institution{JD Explore Academy}\country{ }}

\author{Yuming Li}
\affiliation{\institution{Peking University}\country{ }}

\author{Ruizhe He}
\affiliation{\institution{Zhejiang University}\country{ }}

\author{Haoran Li}
\affiliation{\institution{JD Explore Academy}\country{ }}

\author{Shijia Ge}
\affiliation{\institution{Tsinghua University}\country{ }}

\author{Siming Fu}
\affiliation{\institution{JD Explore Academy}\country{ }}

\begin{abstract}
Reward models guide text-to-image (T2I) systems toward outputs aligned with human preferences. However, typical reward models such as HPSv3 are trained on pre-annotated data from earlier T2I models, without accounting for quality discriminative shifts arising from evolving model capabilities and reinforcement learning (RL) iterations, limiting their broader applicability. In this work, we propose HPSv3++, a reward model framework that elevates the HPSv3 model for varying T2I model capabilities and their RL iteration changes across the full capability-iteration spectrum. Specifically, we first introduce HPDv3++, a 212K dual-dimension preference dataset annotated for text fidelity and aesthetic quality using a recent high-capability (Qwen-Image) model with human supervision. We then propose a two-stage training framework. Stage 1 employs data-aware orthogonal gradient projection to incorporate diverse aesthetic perception from HPDv3++ while preserving the original effective human preference knowledge in HPSv3.
Stage 2 further leverages unlabeled data from T2I models spanning different capability levels and RL iterations, and introduces a joint capability-iterations conditioned signal for the reward model together with a standard deviation-driven unsupervised guidance mechanism, strengthening reward model across the capability-iteration spectrum. HPSv3++ achieves state-of-the-art preference prediction, outperforming HPSv3 9.8\% on HPDv3, 5.5\% on GenAI-Bench, while achieving 79.1\%/88.1\% on our proposed HPDv3++. When used for T2I RL training, it consistently improves GenEval scores across diverse T2I models, demonstrating its wide-range capabilities. The code is available at \url{https://github.com/PlantPotatoOnMoon/HPSv3-PlusPlus}
\end{abstract}


\begin{teaserfigure}
  \centering
  \includegraphics[
    width=\textwidth,
    trim={0 0 0 14mm},
    clip
  ]{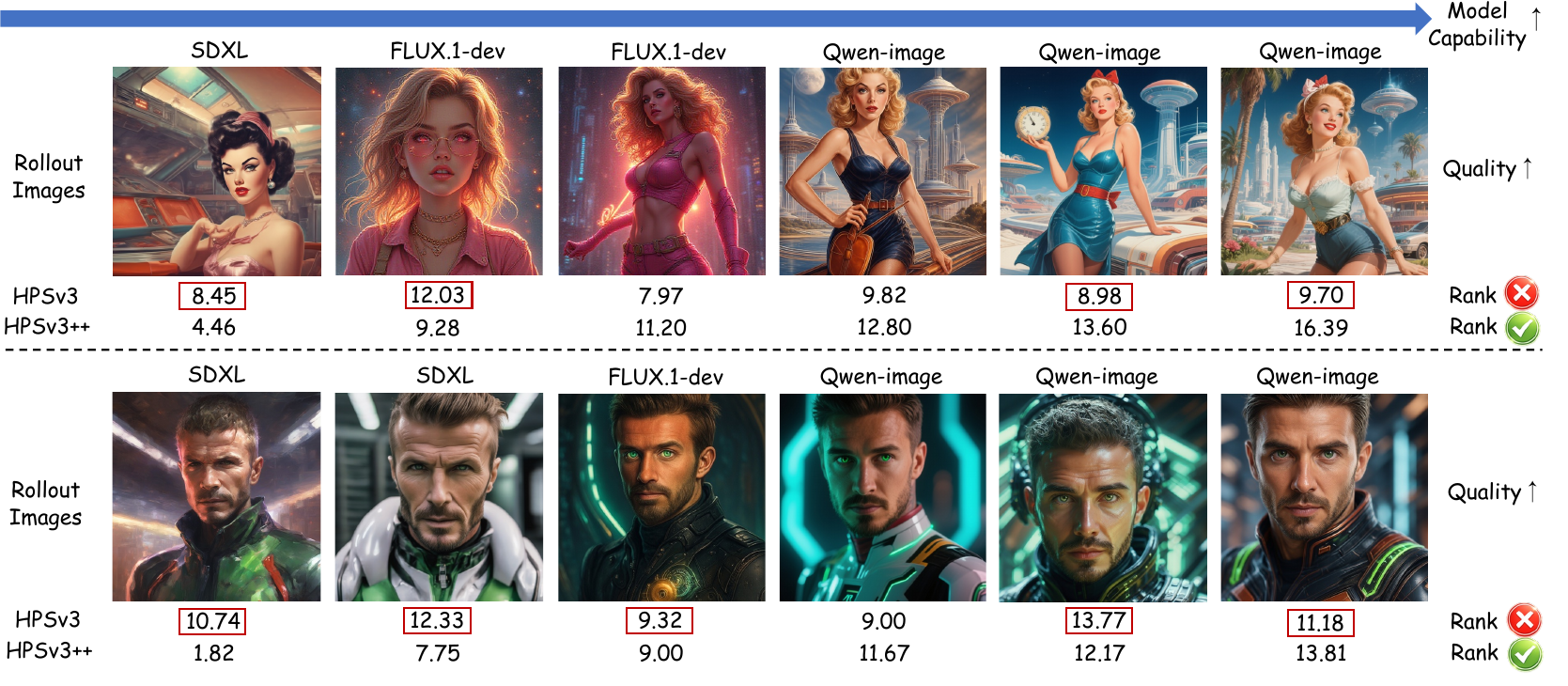}
  \vspace{-4mm}
  \caption{Comparison of reward model scoring on images generated by different T2I models. Images are arranged from low to high quality (left to right). HPSv3 fails to produce a monotonically increasing score sequence, assigning high scores to off-topic or low-quality images while underrating high-quality ones. HPSv3++ produces scores that consistently align with visual quality, enabling reliable ranking across images from T2I models with varying generation capabilities.}
  \Description{A showcase figure displaying generated images from different text-to-image models before and after RL fine-tuning with HPSv3++ reward model.}
  \label{fig:teaser}
\end{teaserfigure}

\maketitle

%% Main paper
\section{Introduction}
\label{sec:intro}
 
\begin{figure}[t]
\centering
\includegraphics[width=\linewidth]{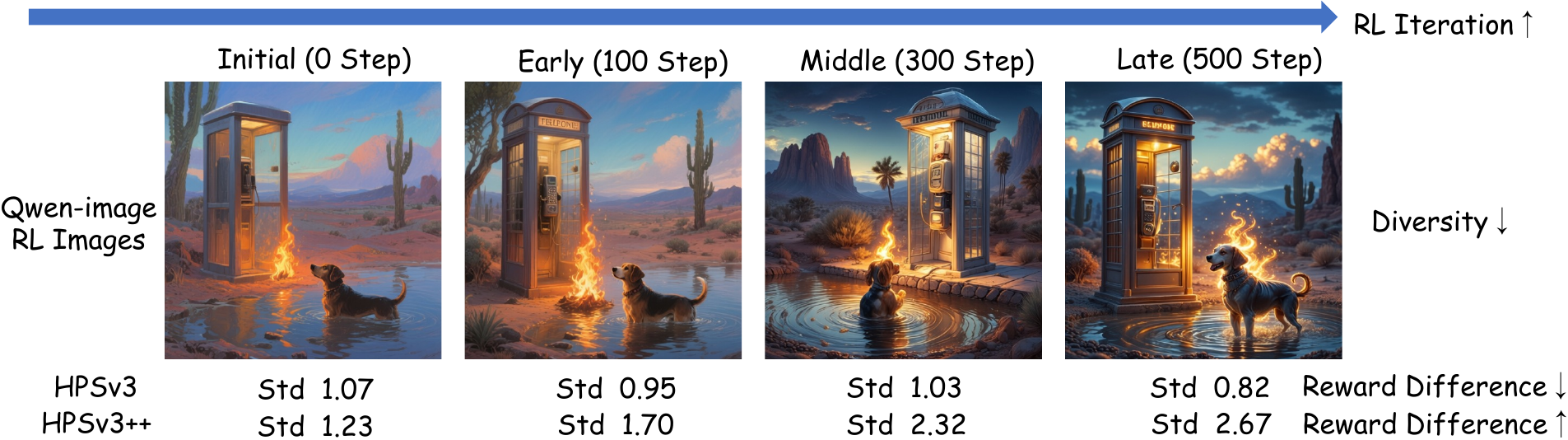}
\caption{Illustration of score standard (std) behavior across RL iterations. As RL progresses, generated images become increasingly similar, reducing diversity. HPSv3 produces decreasing score std (from 1.07 to 0.82), losing the ability to distinguish quality differences. In contrast, HPSv3++ produces monotonically increasing std (from 1.23 to 2.67), maintaining strong quality discriminability across the RL procedure.}
\Description{Comparison of HPSv3 and HPSv3++ score standard deviation across RL iterations on Qwen-Image generated images.}
\label{fig:std_intro}
\end{figure}
 
Reinforcement Learning from Human Feedback (RLHF) has become a key paradigm for improving text-to-image (T2I) generation, enabling models to produce outputs that better align with human preferences in both fidelity and visual quality. At the core of RLHF, reward models (RMs) assign preference-aligned, quality-discriminative scores to images generated by T2I systems, effectively capturing human preferences across varying quality levels. These scores serve as optimization signals that guide T2I models toward higher-quality generations in the reinforcement learning (RL) stage, allowing iterative quality refinement beyond the limits of supervised training alone~\cite{black2024ddpo,fan2024dpok,xue2025dancegrpo}.
A series of RMs, including ImageReward~\cite{xu2024imagereward}, PickScore~\cite{kirstain2023pickapic}, MPS~\cite{he2023mps}, and the advanced HPSv3~\cite{ma2025hpsv3}, have demonstrated strong correlation with human preferences on their respective training distributions.
 
\begin{figure}[t]
\centering
\vspace{-2mm}
\includegraphics[width=\linewidth]{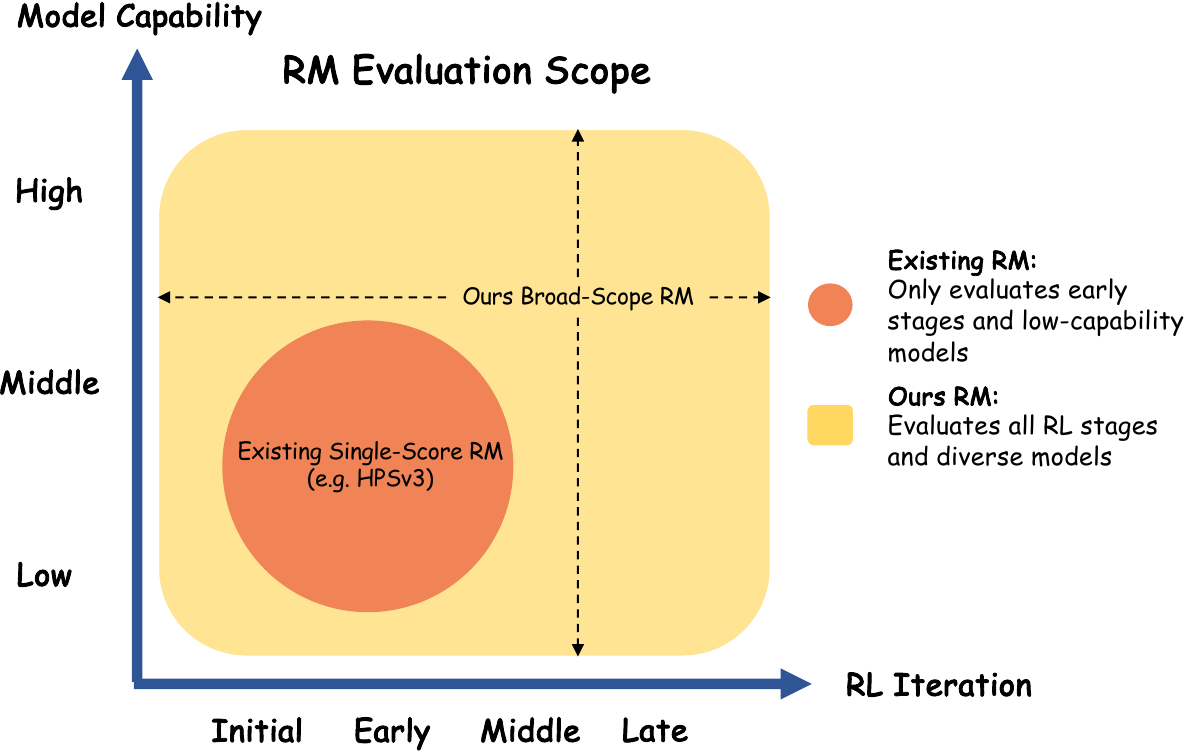}
\vspace{-6mm}
\caption{Comparison of RM evaluation scope. Existing single-score RMs (e.g., HPSv3) is typically effective for low-capability models at early RL iterations (orange region), while our broad-scope RM (HPSv3++) covers the full [\emph{model capability $\times$ RL iteration}] space (yellow region), enabling dynamic calibrated scoring across T2I generation models with diverse capabilities and throughout the RL iterations.}
\vspace{-3mm}
\label{fig:concept}
\end{figure}

However, typical reward models such as HPSv3 are trained on pre-annotated pairwise data from relatively dated T2I models (e.g., SDXL~\cite{podell2024sdxl}), limiting their applicability in broader settings from two aspects. First, as T2I systems have progressed from early diffusion models to frontier models, the visual quality and preference characteristics of generated images have evolved substantially. RMs trained on historical data distributions often struggle to generalize across this expanding capability spectrum, leading to inaccurate preference prediction scores for ranged T2I-generated images (see Fig.~\ref{fig:teaser}).
Second, the RL rollout distribution of the T2I model evolves across RL iterations. A static RM trained on pre-optimization data struggles to meet this distribution shift, results in degraded quality discriminability for the RL rollout across RL iterations (see Fig.~\ref{fig:std_intro}). Together, these limitations confine existing RMs to a narrow region of the [\emph{model capability $\times$ RL iteration}] space (see Fig.~\ref{fig:concept}).

To address these challenges, we introduce HPSv3++, a new RM that not only absorbs human preference knowledge from higher-capability T2I models, but also perceives the T2I model's capability level and RL iteration rollout distribution shift associated with each sample during scoring, achieving dynamic calibration for both dimensions. Therefore, we extend HPSv3++ from the typical static single-score formulation into a conditioned reward modeling framework that covers a broader capability-iteration spectrum.
 
\begin{figure*}[t]
\centering
\includegraphics[width=\linewidth]{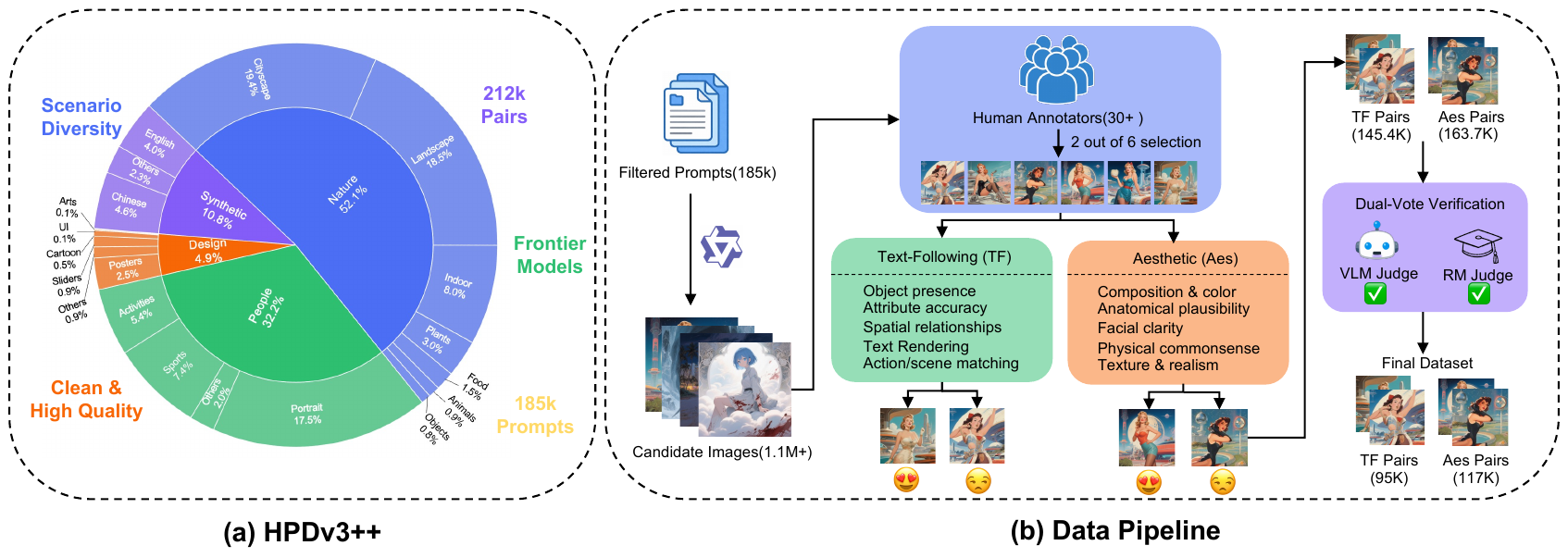}
\caption{Overview of HPDv3++ dataset. (a) Dataset composition: we collect 212K preference pairs across 185K filtered prompts from Qwen-Image, covering diverse scenarios. (b) Data pipeline: we use 185K prompts to generate 1.1M+ candidate images, from which human annotators perform 2-out-of-6 pairwise comparisons along text-following (TF) and aesthetic (Aes) dimensions, yielding 145K and 164K raw pairs respectively. A dual-vote verification step using both a VLM judge and an RM judge removes noisy labels, producing the final 95K text-following and 117K aesthetic pairs.}
\label{fig:data_pipeline}
\end{figure*}

Specifically, we introduce HPDv3++, a 212K dataset built from 185K diverse prompts and images generated by a high-capability T2I model (Qwen-Image~\cite{qwenimage2025}). HPDv3++ is annotated along two complementary dimensions: text-following fidelity and aesthetic quality (see Fig.~\ref{fig:data_pipeline}). To ensure broad coverage, we balance the distribution across more than 20 scenarios. To improve label reliability, human annotations are followed by automatic dual-vote quality control to filter noisy samples. Compared with existing preference datasets, HPDv3++ places greater emphasis on preference supervision under frontier generation model and provides a enhanced supervision basis for reward model training.

Based on HPDv3++, we propose a two-stage training framework that enhances HPSv3, resulting in HPSv3++ (see Fig.~\ref{fig:method}). In Stage 1, inspired by continual learning, we apply data-aware Orthogonal Gradient Descent (OGD)~\cite{lopez2017ogd} on HPDv3++, extending the model's perception to frontier generation models while preserving the original effective human preference knowledge of HPSv3. 
In Stage 2, we perform semi-supervised learning using labeled HPDv3++ and additional unlabeled data from T2I models with varying capabilities and RL iterations.
Specifically, we redesign the RM with a two-dimensional capability–iteration conditioning signal: the T2I model capability is encoded from image features via a Capability Encoder and combined with RL iteration through FiLM conditioning~\cite{perez2018film}. 
Motivated by the the discovery that the RM's quality discriminability is strongly correlated with standard deviation in intra-group of images (see Tab.~\ref{tab:std_motivation}), we further design a standard deviation-driven unsupervised guidance mechanism that increases intra-group score discrimination on unlabeled groups and encourages stronger sensitivity in more challenging capability and RL iteration regions, thereby further improving the model's perception and calibration across both dimensions.
Consequently, HPSv3++ achieves the improved quality discriminative performance on both model capability level (see Fig.~\ref{fig:teaser}) and RL iteration level (see Fig.~
\ref{fig:std_intro}).
 
% Contributions
Our contributions are threefold:
\begin{itemize}
    \item We introduce a 212K dual-dimension preference dataset (HPDv3++), built from 185K prompts and over 1.1M images, annotated for text-following (95K pairs) and aesthetic quality (117K pairs) on a frontier T2I model, with balanced diversity across 20+ scenarios and automated quality control.
    \item We introduce HPSv3++, a two-stage framework for reward model training that improves generalization across model capabilities and adapts to RL-induced distribution shifts. It first applies data-aware OGD for continual learning, then further fine-tunes the RM with capability–iteration conditioned, standard deviation-guided semi-supervised training.
    \item Extensive experiments demonstrating state-of-the-art preference prediction accuracy, achieving 86.7\% on HPDv3 (+9.8\% over HPSv3), 76.3\% on GenAI-Bench (+5.5\%), and 79.1\%/88.1\% on our HPDv3++ aesthetic/text-following benchmarks. When applied as the RL reward signal, diverse generation models consistently improve.
\end{itemize}
 
\section{Related Work}
\label{sec:related}

\subsection{Reward Models for Text-to-Image Generation}
\label{sec:related_rm}

Automatic preference prediction for T2I generation has progressed from generic similarity metrics to dedicated, fine-grained reward models. CLIPScore~\cite{radford2021clip} uses cosine similarity of CLIP embeddings as a lightweight proxy for text-image alignment. ImageReward~\cite{xu2024imagereward} fine-tunes BLIP on human preference annotations to produce a scalar reward correlated with human judgment. PickScore~\cite{kirstain2023pickapic} fine-tunes CLIP on user preference choices from the Pick-a-Pic dataset. HPS~\cite{wu2023hps} and HPSv2~\cite{wu2023hpsv2} fine-tune CLIP on dedicated human preference datasets with progressively broader model coverage. MPS~\cite{he2023mps} introduces multi-perspective scoring that decomposes preference into distinct dimensions. HPSv3~\cite{ma2025hpsv3} adopts a VLM backbone (Qwen2-VL) and trains on a wide-spectrum dataset (HPDv3) covering both legacy and frontier generation models, achieving the strongest reported preference accuracy.

A common thread across all these methods is that they produce a single unconditional score per image, treating all generation models and optimization stages identically. While this design is simple, it fundamentally limits generalization: the same scoring function cannot optimally rank images from vastly different quality distributions. Our work departs from this paradigm by conditioning the reward on model capability and RL iteration, enabling a single model to adapt its behavior across the full spectrum.

\subsection{Reinforcement Learning for Diffusion Models}
\label{sec:related_rl}

Aligning diffusion models with human preference via RL has attracted growing attention and has become a standard practice for improving generation quality beyond supervised training. DDPO~\cite{black2024ddpo} formulates the denoising process as a multi-step Markov decision process and applies policy gradient optimization. DPOK~\cite{fan2024dpok} extends this with KL-regularized policy optimization to stabilize training. DiffusionDPO~\cite{diffusiondpo2024} adapts direct preference optimization to diffusion, bypassing the need for an explicit reward model. REBEL~\cite{rebel2024} introduces a regression-based approach that avoids the high variance of policy gradients. DDRL~\cite{ddrl2024} proposes data-regularized RL using forward KL divergence to anchor the policy to an off-policy data distribution. DanceGRPO~\cite{xue2025dancegrpo} and Flow-GRPO~\cite{flowgrpo2025} unify group relative policy optimization across diffusion models and flow matching models respectively, supporting text-to-image, text-to-video, and image-to-video generation within a single framework. All of these methods rely on a frozen reward model to provide the training signal. As the generation model improves through RL iterations, the output distribution shifts away from the RM's training data, causing score miscalibration and reward hacking. This makes RM quality a critical bottleneck for downstream RL performance, motivating reward models that can adapt to the evolving optimization landscape.

\section{Method}
\label{sec:method}

\begin{figure*}[t]
\centering
\includegraphics[width=0.99\linewidth]{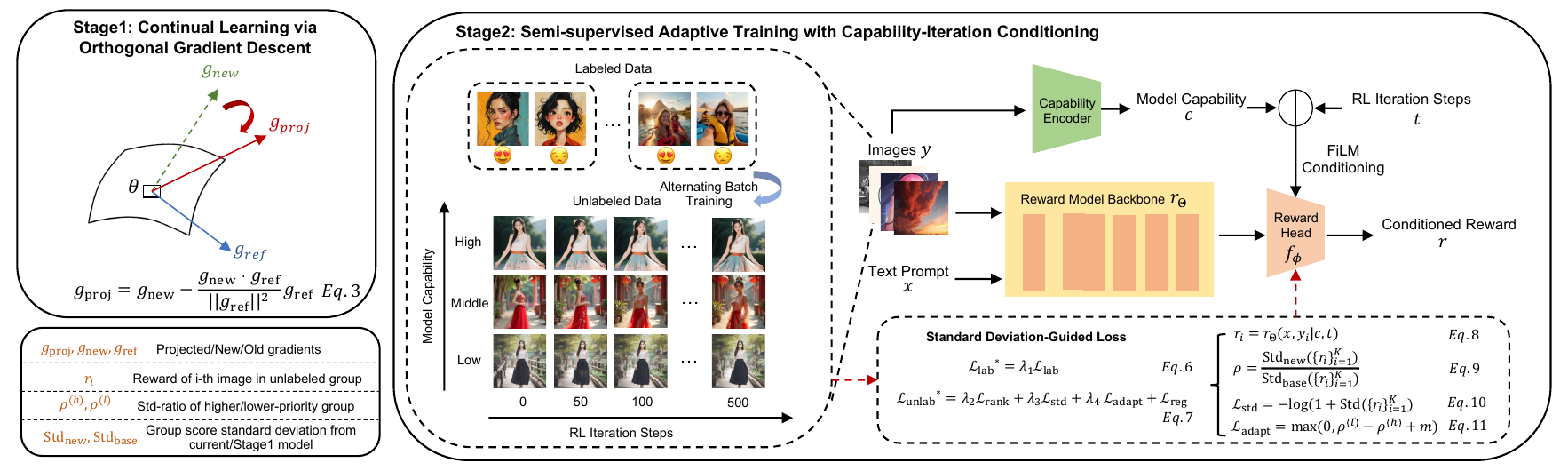}
\caption{Overview of the HPSv3++ training framework. Stage~1 (top): Inspired by continual learning, we train HPSv3++ using Orthogonal Gradient Descent that projects the new-data (HPDv3++) gradient onto the orthogonal complement of the reference-data (HPDv3) gradient to keep HPSv3's original knowledge while learning new data's preferences. Stage~2 (bottom): (Left) Semi-supervised Adaptive Training combines labeled pairwise data (HPDv3) and unlabeled rollout data from various T2I models on RL training, which adopts alternating-batch training with intra-group std maximization and adaptive ratio constraints; (Right) Capability-Iteration conditioning infers model capability implicitly from image features by Capability Encoder and receives the RL iteration as the condition explicitly, jointly modulating the reward through FiLM conditioning.}
\label{fig:method}
\end{figure*}

\subsection{Overview}
\label{sec:method_overview}

Our goal is to build a reward model that remains reliable across both diverse generation model capabilities and evolving RL iteration stages. To this end, HPSv3++ consists of three key components. First, we construct HPDv3++ (see Fig.~\ref{fig:data_pipeline}), a dual-dimension preference dataset that broadens preference supervision toward frontier high-capability generation model distributions (Sec.~\ref{sec:method_dataset}). Second, as shown in the left of Fig.~\ref{fig:method}, we perform Stage~1 continual learning with data-aware Orthogonal Gradient Descent (OGD) to extend the reward model to the new data distribution while preserving the original preference knowledge inherited from HPSv3 (Sec.~\ref{sec:method_ogd}). Third, as depicted in the right of Fig.~\ref{fig:method}, we perform Stage~2 semi-supervised adaptive training with capability- and iteration-conditioned reward modeling, using both labeled preference data and unlabeled rollout data to improve calibration across the capability-iteration space (Sec.~\ref{sec:method_semisup}). Overall, this design transforms HPSv3 from a static unconditional reward model into a capability- and iteration-aware reward model, enabling more robust preference prediction and more reliable reward supervision for downstream RLHF.

\subsection{HPDv3++: Dual-Dimension Preference Dataset}
\label{sec:method_dataset}

\paragraph{Motivation and Overview.}
A key motivation of HPSv3++ is to broaden the coverage of existing preference datasets, particularly by supplementing preference supervision under distributions induced by frontier high-capability generation models. To this end, we introduce \textbf{HPDv3++}, a large-scale preference dataset that extends HPDv3~\cite{ma2025hpsv3} along two axes: model coverage and annotation granularity. In contrast to many existing preference datasets, which construct preference pairs across different generation models, HPDv3++ is annotated entirely on multiple rollouts from the same frontier generator for each prompt. This design better matches the downstream RL setting, where preference comparison is performed among images generated by the same model, and avoids over-reliance on pairs with large capability gaps that may otherwise encourage reward hacking. As shown in Fig.~\ref{fig:data_pipeline}, built from 185K diverse prompts spanning 20+ categories, the dataset is designed to provide a richer and more fine-grained supervision basis for preference modeling under frontier generation model distributions.

\paragraph{Image Generation and Prompt Curation.}
Specifically, we collect approximately 185K diverse prompts and generate six images per prompt using Qwen-Image, a recent high-capability generation model that is absent from the training set of HPDv3. This yields over 1.1M candidate images. By introducing Qwen-Image into the data construction pipeline, we effectively expand the coverage of existing datasets toward frontier generation model distributions. To ensure prompt quality, we further employ Qwen3-VL-32B-Instruct to filter out noisy, ambiguous, or uninstructive prompts, retaining only those that provide meaningful evaluation signals.

\paragraph{Dual-Dimension Human Annotation.}
Unlike prior datasets that collect only a single overall preference label, HPDv3++ annotates image pairs along two complementary dimensions. The first is \emph{text-following fidelity}, which measures how faithfully an image reflects the prompt semantics, including object presence, attribute accuracy, spatial relationships, and text rendering. The second is \emph{aesthetic quality}, which goes beyond composition and color harmony to cover a broader range of perceptual criteria, such as anatomical plausibility, facial clarity, physical commonsense, texture realism, and the absence of obvious AI-generated artifacts. For each prompt, annotators compare pairs sampled from the six candidate images and provide a preference judgment along both dimensions. This finer-grained annotation scheme disentangles two fundamental yet often conflicting aspects of image quality, thereby providing richer supervision for reward model training and enabling dimension-specific analysis of model behavior. To ensure annotation quality, we also establish a two-stage manual quality inspection process. In the first stage, annotation group leaders inspect the annotation results. In the second stage, technical staff conduct additional spot checks through sampling, further improving the overall annotation quality.

\paragraph{Automatic Quality Assurance and Dataset Comparison.}
After human annotation, we further design an automatic quality assurance procedure to identify and remove unreliable labels. Specifically, Qwen3-VL-32B-Instruct, serving as a VLM judge, and the existing HPSv3 model independently evaluate each annotated image pair. A sample is discarded only when both automated judges disagree with the human annotation. This conservative criterion preserves the majority of human labels while effectively filtering out obvious errors. The final dataset contains approximately 117K aesthetic pairs and 95K text-following pairs, for a total of around 212K pairs. The aesthetic dimension contains 111,650 training pairs and 5,720 test pairs, and the text-following dimension contains 91,003 training pairs and 4,465 test pairs. Compared with existing datasets such as ImageRewardDB~\cite{xu2024imagereward}, Pick-a-Pic~\cite{kirstain2023pickapic}, HPDv2~\cite{wu2023hpsv2}, and HPDv3~\cite{ma2025hpsv3}, HPDv3++ is the first dataset to provide dual-dimension annotations covering both text-following fidelity and aesthetic quality, while specifically supplementing preference supervision under frontier high-capability generation model distributions.

\subsection{Stage 1: Continual Learning via Orthogonal Gradient Descent}
\label{sec:method_ogd}

Stage~1 aims to extend the reward model's preference modeling capability to the frontier generation model distributions covered by HPDv3++ while preserving its original preference prediction capability learned from HPDv3. Direct fine-tuning on the new dataset may lead to catastrophic forgetting of previously acquired knowledge. To address this, we combine backbone migration with orthogonal gradient projection to enable continual learning under the new distribution.

\paragraph{Task Setting and Reward Formulation.}
Following HPSv3, we formulate reward prediction as a backbone-head composition. Given a text prompt $x$ and an image $y$, the reward model first extracts a joint multimodal representation through the backbone $E_{\theta}$, and then maps it to a scalar reward score through the reward head $f_{\phi}$. The Stage~1 reward function is defined as
\begin{equation}
r_{\Theta}(x,y)=f_{\phi}(E_{\theta}(x,y)).
\end{equation}
For labeled preference data, given a preferred/dispreferred image pair $(y_w,y_l)$ under the same prompt $x$, we optimize the supervised pairwise preference loss
\begin{equation}
\mathcal{L}_{\text{lab}}
=
-\log \sigma\big(r_{\Theta}(x,y_w)-r_{\Theta}(x,y_l)\big).
\end{equation}
Stage~1 follows this unconditional reward formulation and aims to extend the original HPSv3 reward model to the frontier generation model distributions covered by HPDv3++ while preserving its original preference modeling capability.

\paragraph{Model Architecture.}
We migrate the HPSv3 backbone from Qwen2-VL-7B~\cite{wang2024qwen2vl} to Qwen3-VL-8B~\cite{bai2025qwen3vltechnicalreport} to leverage its stronger visual representation capability and larger hidden dimension. The reward head adopts a three-layer MLP with a RankNet architecture, mapping the pooled hidden state to a scalar reward score. We initialize the model from an HPSv3 variant already adapted to the Qwen3-VL backbone, and train the full model on HPDv3++ with the supervised preference loss above, including both the vision encoder and the language model.

\paragraph{Orthogonal Gradient Descent.}
To mitigate forgetting in continual learning, we adopt Orthogonal Gradient Descent (OGD)~\cite{lopez2017ogd}, which constrains parameter updates to directions that do not conflict with the original task. At each training step, we compute the new gradient $\mathbf{g}_{\text{new}}$ from a batch of HPDv3++ data and the reference gradient $\mathbf{g}_{\text{ref}}$ from a batch of HPDv3 reference data, and determine their compatibility through the inner product between them. When $\mathbf{g}_{\text{new}} \cdot \mathbf{g}_{\text{ref}} > 0$, the two gradients are considered compatible, and $\mathbf{g}_{\text{new}}$ is applied directly. When $\mathbf{g}_{\text{new}} \cdot \mathbf{g}_{\text{ref}} \leq 0$, we project $\mathbf{g}_{\text{new}}$ onto the orthogonal complement of $\mathbf{g}_{\text{ref}}$:
\begin{equation}
\mathbf{g}_{\text{proj}}
=
\mathbf{g}_{\text{new}}
-
\frac{\mathbf{g}_{\text{new}}\cdot\mathbf{g}_{\text{ref}}}{\|\mathbf{g}_{\text{ref}}\|^2}\,\mathbf{g}_{\text{ref}}.
\end{equation}
This operation removes the component of the new gradient that conflicts with the reference gradient, allowing the model to absorb new knowledge while preserving existing capabilities. The projection is computed at the parameter level, and the additional overhead is negligible since each step only requires one extra forward-backward pass on the reference batch. After Stage~1, the model acquires preference modeling capability over frontier generation model distributions while maintaining its original generalization performance, providing a strong initialization for the subsequent capability-aware training stage.

\begin{table}[t]
\centering
\caption{Per-group score std of the Stage~0 (original HPSv3) and Stage~1 (OGD) models on seen and unseen distributions (500 groups each).}
\label{tab:std_motivation}
\small
\begin{tabular}{lccc}
\toprule
Distribution & Seen/Unseen & Stage 0 Std & Stage 1 Std \\
\midrule
Qwen-Image & Unseen & 2.79 & \textbf{5.63} \small{(+102\%)} \\
SD 1.5 & Seen & 3.53 & 3.65 \small{(+3.4\%)} \\
\bottomrule
\end{tabular}
\end{table}

\begin{table*}[t]
\centering
\caption{Pairwise preference prediction accuracy (\%) on multiple benchmarks. Best results are in bold, and second-best results are underlined. ``---'' indicates that the result is not reported by the official HPSv3 model on our HPDv3++ benchmarks.}
\label{tab:main_results}
\small
\begin{tabular}{l|cccccc|cc}
\toprule
\multirow{2}{*}{Model} & \multirow{2}{*}{ImageReward} & \multirow{2}{*}{PickScore} & \multirow{2}{*}{HPDv3} & \multirow{2}{*}{GenAI-Bench} & \multicolumn{2}{c|}{MJ-Bench} & \multicolumn{2}{c}{HPDv3++ (Ours)} \\
 & & & & & Align. & Qual. & Aesth. & T-Fol. \\
\midrule
CLIP ViT-H/14   & 57.1 & 60.8 & 48.6 & 56.0 & 58.4 & 68.4 & 51.8 & 55.5 \\
Aesthetic Score & 57.4 & 56.8 & 59.9 & 57.3 & 56.9 & 83.0 & 56.1 & 57.5 \\
ImageReward     & 65.1 & 61.1 & 58.6 & 63.4 & 64.2 & 81.8 & 58.5 & 63.6 \\
PickScore       & 61.6 & \underline{70.5} & 65.6 & 70.0 & 65.0 & 89.6 & 57.9 & 63.1 \\
\midrule
UnifiedReward   & 63.8 & 62.5 & 72.0 & \underline{72.4} & 60.8 & 97.2 & 57.9 & 67.1 \\
HPSv3           & \textbf{66.8} & \textbf{72.2} & \underline{76.9} & 70.8 & \underline{69.0} & \underline{97.6} & \underline{74.8} & \underline{73.3} \\
\midrule
\textbf{HPSv3++ (Ours)} & \underline{66.0} & \underline{70.5} & \textbf{86.7} & \textbf{76.3} & \textbf{69.8} & \textbf{98.7} & \textbf{79.1} & \textbf{88.1} \\
\bottomrule
\end{tabular}
\end{table*}

\subsection{Stage~2: Semi-supervised Adaptive Training with Capability-Iteration Conditioning}
\label{sec:method_semisup}

Stage~1 extends the reward model to frontier generation model distributions, but its output remains unconditional and cannot explicitly adapt to distribution shifts across different model capability levels and RL iteration stages. To address this, Stage~2 introduces capability- and iteration-conditioned reward modeling and performs semi-supervised adaptive training with both labeled preference data and unlabeled rollout data, enabling the model to produce more stable and calibrated rewards over a broader capability-iteration space.

\paragraph{Motivation.}
We observe that the intra-group score standard deviation of a Bradley--Terry reward model is closely related to distribution familiarity. As shown in Table~\ref{tab:std_motivation}, the model typically exhibits higher intra-group score std on seen distributions and lower std on unseen ones; after Stage~1 OGD training, the std on newly covered distributions increases substantially, while the std on previously covered distributions changes only slightly. This suggests that score std serves as a reliable indicator of distribution familiarity. We further validate this observation on new unlabeled rollout data: maximizing intra-group score std maintains the Stage~1 model's HPDv3 accuracy at approximately 77.5\% throughout 3K training steps, whereas minimizing std degrades accuracy from 77.3\% to 75.5\%. Motivated by these findings, we introduce a std-guided adaptive objective in Stage~2.

\paragraph{Unlabeled Rollout Data Construction.}
To learn image distributions across different capability levels and RL iteration stages, we construct a large-scale unlabeled rollout dataset involving multiple diffusion backbones, including SD 1.5~\cite{rombach2022sd}, SDXL~\cite{podell2024sdxl}, FLUX.1-dev~\cite{flux2024}, and Qwen-Image~\cite{qwenimage2025}. Along the capability axis, we form multi-model image groups for the same prompt using generators of different quality levels. Along the iteration axis, we sample images for the same set of prompts from checkpoints at multiple RL steps during GRPO training, where the Stage~1 model serves as the reward signal. In total, we obtain approximately 12K multi-model prompts and 3K multi-iteration prompts, providing dense coverage over both axes without requiring any human annotation.

\paragraph{Capability-Iteration Conditioned Reward Formulation.}
Built upon the Stage~1 reward model, we extend the unconditional reward to a capability- and iteration-conditioned form:
\begin{equation}
r_{\Theta}(x,y\mid c,t)=f_{\phi}(E_{\theta}(x,y),c,t),
\end{equation}
where $c$ denotes the generator capability condition and $t$ denotes the normalized RL iteration condition. Here, $c$ is encoded and learned from image features by the Capability Encoder, while $t$ is obtained by normalizing the RL iteration steps from downstream diffusion models. The two conditions are then injected into the reward head through FiLM conditioning~\cite{perez2018film}, enabling a single reward model to output calibrated scores for different capability levels and RL iteration stages.

\paragraph{Standard Deviation-Guided Loss.}
Motivated by the above observations, we propose a standard deviation-guided semi-supervised adaptive training strategy that uses unlabeled rollout data to improve the model's discriminative ability over the capability--iteration space. To avoid drifting away from human preference supervision during unlabeled adaptation, we jointly use aesthetic subset of HPDv3++ labeled data and unlabeled rollout data and optimize them with alternating-batch training.

For labeled data, we continue to use the supervised preference loss $\mathcal{L}_{\text{lab}}$ defined in Stage~1. For unlabeled rollout data, we optimize
\begin{equation}
\mathcal{L}
=
\begin{cases}
\mathcal{L}_{\text{lab}}^{*}, & \text{for labeled batches},\\[4pt]
\mathcal{L}_{\text{unlab}}^{*}, & \text{for unlabeled batches},
\end{cases}
\end{equation}
with
\begin{equation}
\mathcal{L}_{\text{lab}}^{*}
=
\lambda_{1}\mathcal{L}_{\text{lab}},
\end{equation}
\begin{equation}
\mathcal{L}_{\text{unlab}}^{*}
=
\lambda_{2}\mathcal{L}_{\text{rank}}
+
\lambda_{3}\mathcal{L}_{\text{std}}
+
\lambda_{4}\mathcal{L}_{\text{adapt}}
+
\mathcal{L}_{\text{reg}}.
\end{equation}
where
\begin{equation}
r_i = r_{\Theta}(x,y_i \mid c,t),
\end{equation}
\begin{equation}
\rho
=
\frac{\operatorname{Std}_{\text{new}}\big(\{r_i\}_{i=1}^{K}\big)}
{\operatorname{Std}_{\text{base}}\big(\{r_i\}_{i=1}^{K}\big)},
\end{equation}
\begin{equation}
\mathcal{L}_{\text{std}}
=
-\log\!\Big(1+\operatorname{Std}\big(\{r_i\}_{i=1}^{K}\big)\Big),
\end{equation}
\begin{equation}
\mathcal{L}_{\text{adapt}}
=
\max\!\big(0,\rho^{(\ell)}-\rho^{(h)}+m\big).
\end{equation}

Here, $r_i$ denotes the conditioned reward of the $i$-th image in an unlabeled image group under prompt $x$, and $\operatorname{Std}_{\text{base}}$ denotes the intra-group score std produced by the Stage~1 model. $\mathcal{L}_{\text{rank}}$ enforces relative ordering across rollout groups and preserves the model's ability to distinguish quality differences under unlabeled adaptation. $\mathcal{L}_{\text{std}}$ increases intra-group score dispersion, encouraging the reward model to remain sensitive to subtle quality variations within the same prompt, while the logarithmic form prevents the score spread from growing without bound. $\mathcal{L}_{\text{adapt}}$ compares the std ratios of a higher-priority group $(h)$ and a lower-priority group $(\ell)$, where priority is determined jointly by model capability and RL iteration, and thus explicitly encourages stronger generators or later RL stages to exhibit larger relative std gains. Finally, $\mathcal{L}_{\text{reg}}$ collects auxiliary stabilization terms, including std-ratio constraints, predicted-condition supervision, and a weak reward-magnitude regularizer, which together prevent the adaptive objectives from driving the score distribution to unstable extremes. By combining these objectives, Stage~2 improves both discrimination and calibration over the capability--iteration space, while maintaining alignment with human preference supervision. Detailed implementations are provided in the appendix.
\section{Experiments}
\label{sec:exp}
 
\begin{figure}[t]
\centering
\includegraphics[width=\linewidth]{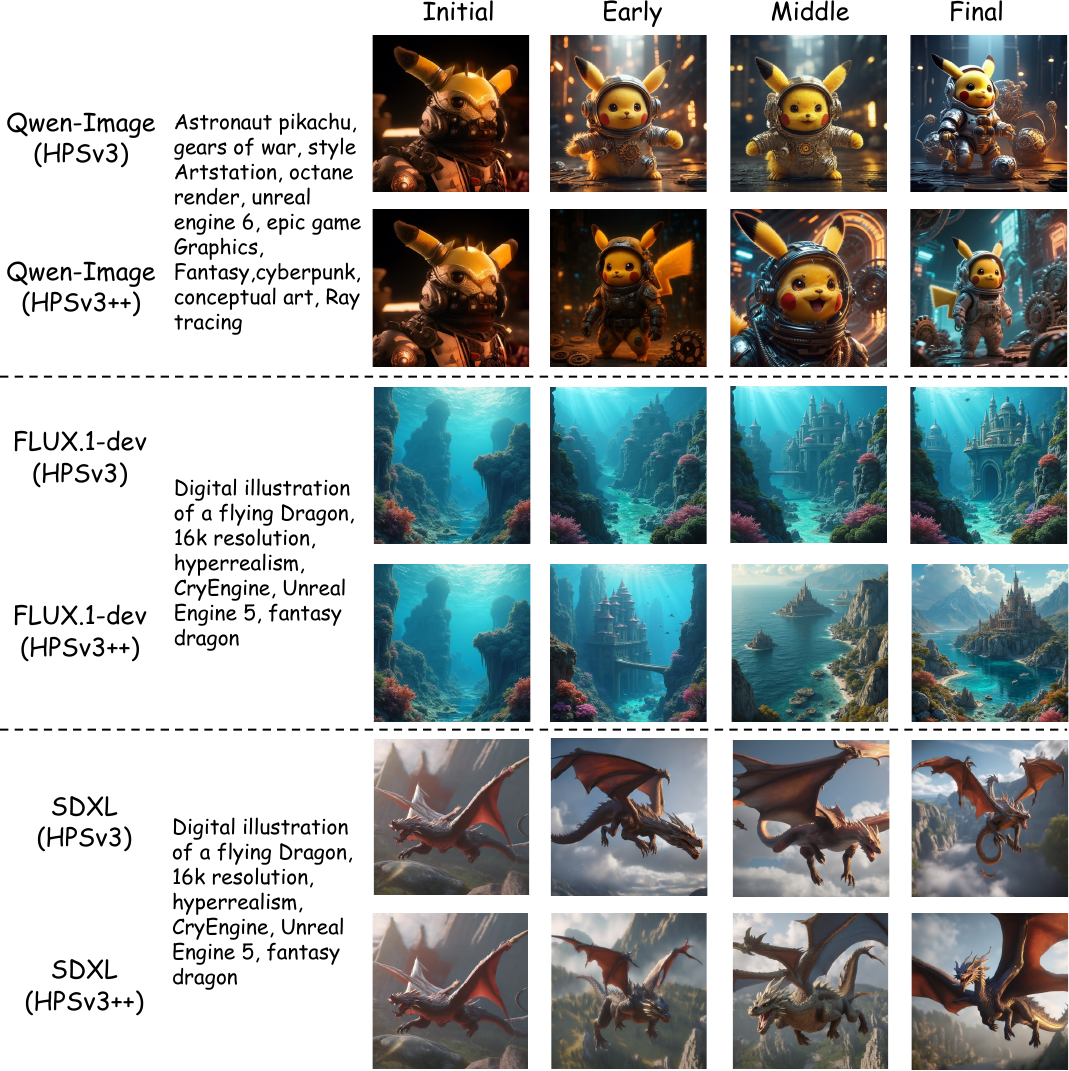}
\vspace{-8mm}
\caption{Qualitative comparison of RL fine-tuning using HPSv3 vs.\ HPSv3++ as reward models across three generation backbones (Qwen-Image, FLUX.1-dev, SDXL). For each model, the top row shows results guided by HPSv3 and the bottom row by HPSv3++. Columns correspond to RL stages from initial to final. HPSv3++ produces visually superior outputs with more consistent quality progression across all backbones and RL iterations.}
\Description{A grid of generated images comparing RL fine-tuning progression using HPSv3 and HPSv3++ as reward models for Qwen-Image, FLUX.1-dev, and SDXL.}
\label{fig:rl_showcase}
\end{figure}
 
\subsection{Preference Prediction Accuracy}
\label{sec:exp_accuracy}
 
\paragraph{Setup.}
Our reward model adopts Qwen3-VL-8B~\cite{bai2025qwen3vltechnicalreport} as the backbone and uses a three-layer RankNet MLP as the reward head. Training follows a two-stage scheme. In Stage 1, we train on 191,466 HPDv3++ pairwise samples across the aesthetic and text-following dimensions, while using the HPDv3 training set (285K pairs) as the OGD reference data, and the full model is trained for 1 epoch. In Stage 2, we freeze the vision encoder and language model, and train only the visual merger, reward head, and a randomly initialized Capability Encoder for approximately 2 epochs. The labeled data consists of 111,650 aesthetic pairs from HPDv3++, while the unlabeled rollout data includes approximately 12K multi-model prompts and 3K multi-iteration prompts collected from generators of different capability levels, including SD1.5, SDXL, FLUX.1-dev, and Qwen-Image. Additional training and implementation details are provided in the appendix~\ref{sec:appendix_training_details}.
 
\paragraph{Results.}
Table~\ref{tab:main_results} summarizes the pairwise preference prediction accuracy on multiple benchmarks. HPSv3++ performs best on benchmarks that better reflect frontier generation model distributions, achieving 76.3\% on GenAI-Bench, 86.7\% on HPDv3, and 69.8\% / 98.7\% on the alignment / quality subsets of MJ-Bench, outperforming HPSv3 by 5.5, 9.8, 0.8, and 1.1 percentage points, respectively. On earlier benchmarks such as ImageReward and PickScore, it remains competitive, obtaining 66.0\% and 70.5\%. On our HPDv3++ test set, HPSv3++ achieves 79.1\% and 88.1\% on the aesthetic and text-following splits, respectively, further validating its strong discriminative ability on outputs from frontier generation models. Additional user study results are provided in the supplementary material.
\begin{figure}[t]
\centering
\includegraphics[width=\linewidth]{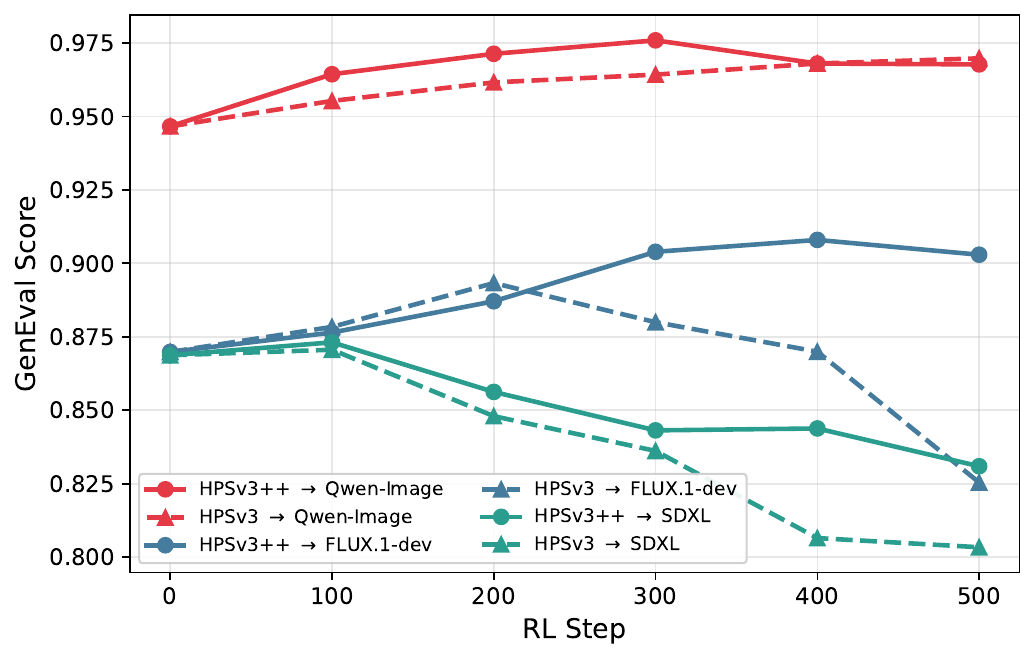}
\caption{GenEval scores during RL fine-tuning (0--500 steps) using HPSv3++ (solid) and HPSv3 (dashed) as reward models across three generation backbones.}
\Description{Line chart showing GenEval scores over RL steps for six settings: HPSv3++ and HPSv3 as reward models, each applied to Qwen-Image, FLUX.1-dev, and SDXL.}
\label{fig:rl_geneval}
\end{figure}

%%%%%%%%%%%%%%%%%%%%%%%%%%%%%%%%%%%%%%%%%%%%%%%%%%%%%%%%%%%%

\subsection{Reward Model for RL across Different Generation Models}
\label{sec:exp_rl}
 
\paragraph{Setup.}
To evaluate whether HPSv3++ provides effective reward signals for downstream RL fine-tuning, we apply Flow-GRPO~\cite{flowgrpo2025} to three text-to-image backbones with different capability levels, namely SDXL~\cite{podell2024sdxl}, FLUX.1-dev~\cite{flux2024}, and Qwen-Image~\cite{qwenimage2025}. For each backbone, we compare HPSv3++ against HPSv3 as the reward model under the same RL configuration, and evaluate checkpoints at steps 0, 100, 200, 300, 400, and 500 using HPSv2, Aesthetic Score, CLIPScore, and GenEval~\cite{geneval2024}. Detailed experimental settings and the full quantitative results are provided in the appendix~\ref{sec:appendix_more_rl_results}.
 
\paragraph{Results.}
As shown in Fig.~\ref{fig:rl_geneval}, HPSv3++ yields overall stronger and more stable RL behavior than HPSv3 across generation backbones with different model capabilities and across different stages of RL optimization. In particular, it achieves clearer improvements on FLUX.1-dev and SDXL, indicating that the proposed reward model provides more effective and robust optimization signals across model capabilities. On Qwen-Image, HPSv3++ also delivers competitive RL performance and stronger improvements during optimization. These results suggest that capability- and iteration-aware reward modeling is beneficial for downstream RL across diverse generation models, rather than being restricted to a single backbone or capability regime. Full metric comparisons are reported in Appendix Table~\ref{tab:rl_all_results}, and a controlled qualitative comparison on Qwen-Image is provided in Appendix Fig.~\ref{fig:rl_qual_compare}.

%%%%%%%%%%%%%%%%%%%%%%%%%%%%%%%%%%%%%%%%%%%%%%%%%%%%%%%%%%%%

\subsection{User Study}
\label{sec:appendix_user_study}
 
To further assess whether HPSv3++ is better aligned with human preference in downstream RL, we conduct a user study on Qwen-Image rollout results. Specifically, we compare Qwen-Image models trained with the same Flow-GRPO-based optimization setup, differing only in the reward model used during optimization, namely HPSv3 and HPSv3++~\cite{ma2025hpsv3,flowgrpo2025,qwenimage2025}. The evaluation is conducted on 100 prompts with 12 human annotators. For each prompt, annotators are presented with the two generated images and asked to select the one that better matches the prompt and exhibits better overall visual quality. Representative comparison examples are shown in Fig.~\ref{fig:rl_qual_compare}.
 
The final win rate is computed as the percentage of comparisons in which the image generated with HPSv3++-based optimization is preferred over that generated with HPSv3. As summarized in Table~\ref{tab:user_study}, HPSv3++ achieves a 77.5\% human win rate against HPSv3, further supporting that it provides reward signals that are better aligned with human preference in downstream RL.
 
\begin{table}[t]
\centering
\caption{Human evaluation on 100 prompts using Qwen-Image rollout results. Both models are trained under identical Flow-GRPO hyperparameters, differing only in the reward model used during optimization. Higher is better.}
\label{tab:user_study}
\small
\begin{tabular}{lccc}
\toprule
Compared Models & \#Prompts & HPSv3++ Win Rate & HPSv3 Win Rate \\
\midrule
HPSv3++ vs. HPSv3 & 100 & 77.5\% & 22.5\% \\
\bottomrule
\end{tabular}
\end{table}

%%%%%%%%%%%%%%%%%%%%%%%%%%%%%%%%%%%%%%%%%%%%%%%%%%%%%%%%%%%%

\subsection{Ablation Studies}
\label{sec:exp_ablation}
 
\begin{table}[t]
\centering
\caption{Ablation studies on key design choices (\%).}
\label{tab:ablation}
\small
\begin{tabular}{lccc}
\toprule
Variant & HPDv3 & Aesth. & T-Fol. \\
\midrule
HPSv3 (Qwen2-VL) & 76.9 & 74.5 & 73.3 \\
HPSv3 (Qwen3-VL) & 77.6 & 67.2 & 67.4 \\
+ Stage 1 (OGD) & 76.5 & 74.7 & 79.8 \\
+ Stage 1 \& 2 (Full) & \textbf{86.7} & \textbf{79.1} & \textbf{88.1} \\
\midrule
CapEncoder, pretrained + frozen & 77.8 & 75.1 & 80.9 \\
CapEncoder, pretrained + trainable & 81.8 & 77.1 & 85.3 \\
CapEncoder, random init + trainable & \textbf{86.7} & \textbf{79.1} & \textbf{88.1} \\
\bottomrule
\end{tabular}
\end{table}
 
We conduct ablation studies to analyze the contribution of key design choices. All variants use the same OGD model as initialization and are evaluated on HPDv3 and our HPDv3++ aesthetic and text-following test sets. Results are shown in Table~\ref{tab:ablation}.
 
\paragraph{Effect of the two-stage training.}
To verify the necessity of both training stages, we compare the full HPSv3++ model with a variant that uses only the Stage 1 OGD-trained model. The Stage 1 model has absorbed knowledge from HPDv3++ through continual learning but has not been exposed to multi-model rollout data, nor does it explicitly model capability- and RL-iteration-dependent score adaptation. Removing Stage 2 leads to a consistent and substantial drop across all three benchmarks: HPDv3 decreases from 86.7\% to 76.5\%, aesthetic accuracy from 79.1\% to 74.7\%, and text-following accuracy from 88.1\% to 79.8\%. This demonstrates that OGD alone, while effective at preventing forgetting, is insufficient for learning the capability- and iteration-aware scoring behavior required for robust reward modeling across heterogeneous generation models and RL stages. Stage 2 provides the additional adaptive supervision needed to establish more reliable ranking and calibration behavior in downstream RL.
 
\paragraph{Capability Encoder strategy.}
We investigate three strategies for the Capability Encoder, which is responsible for implicitly inferring model capability from visual features. In the first variant (v4a), we pretrain the Capability Encoder on source-model prediction and freeze it during Stage 2 training. This setting achieves 77.8\% on HPDv3, 75.1\% on aesthetics, and 80.9\% on text-following, only slightly above the Stage 1 baseline. The frozen encoder cannot adapt to the rollout distributions encountered in Stage 2, which limits the model's ability to exploit capability information effectively. In the second variant (v4b), we initialize the Capability Encoder from the same pretrained model but allow it to be fine-tuned jointly with the rest of Stage 2. This improves performance to 81.8\% on HPDv3, 77.1\% on aesthetics, and 85.3\% on text-following, confirming that end-to-end adaptation is important. In the final variant (v4c), we instead initialize the Capability Encoder randomly and optimize it jointly during Stage 2. This yields the best results, reaching 86.7\%, 79.1\%, and 88.1\% on the three benchmarks, respectively. These results suggest that, for our final setting, learning the capability representation directly from the Stage 2 objective is more effective than relying on a pretrained initialization. Overall, the key factor is not merely introducing a Capability Encoder, but training it in a way that is tightly coupled to the semi-supervised adaptation objective.

%%%%%%%%%%%%%%%%%%%%%%%%%%%%%%%%%%%%%%%%%%%

\subsection{Iteration-Aware Score Std Analysis}
\label{sec:exp_std}
 
\begin{figure}[t]
\centering
\includegraphics[width=\linewidth]{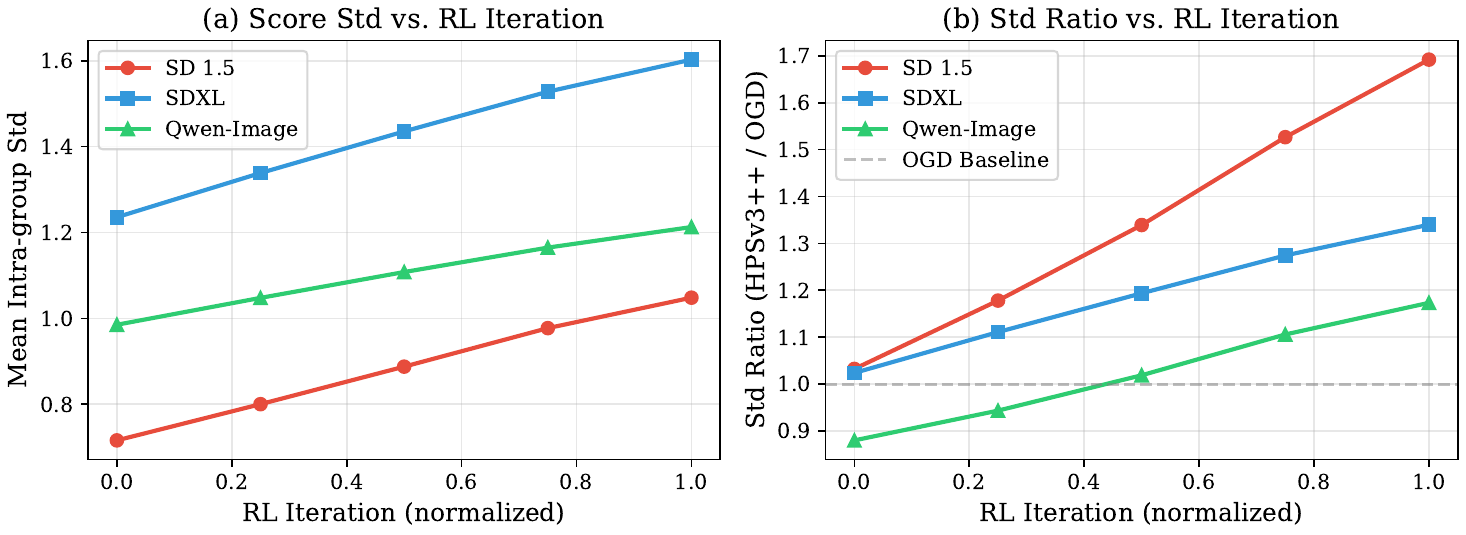}
\caption{Score std analysis of HPSv3++. (a) Mean intra-group std increases monotonically with RL iteration input across all generation model tiers. (b) Std ratio (HPSv3++ / Stage~1 baseline) starts near 1.0 at iter=0 and grows with iteration, confirming that FiLM conditioning adaptively modulates scoring sensitivity. The dashed line indicates the Stage~1 baseline level.}
\label{fig:std_analysis}
\end{figure}

To verify that the conditioned reward model responds correctly to the explicit RL iteration input, we fix the input images and vary only the normalized iteration value. As shown in Fig.~\ref{fig:std_analysis}, the mean intra-group score std increases monotonically with iteration across all generation model tiers. Specifically, SD~1.5 increases from 0.72 at iter=0 to 1.05 at iter=1.0, SDXL from 1.24 to 1.60, and Qwen-Image from 0.99 to 1.21, with 98.3\% of test groups exhibiting strictly increasing std. Moreover, the std ratio relative to the Stage~1 baseline remains close to 1.0 at iter=0 and increases with iteration, reaching approximately 1.4 for SDXL and Qwen-Image at iter=1.0, while 90.5\% of ratios stay within the stable range of $[0.8, 1.5]$. These results confirm that FiLM conditioning effectively modulates scoring sensitivity according to the RL stage, while the ratio constraint prevents both score collapse and unbounded inflation.
 
\section{Conclusion}
\label{sec:conclusion}
We present HPSv3++, a capability- and iteration-aware reward model for text-to-image generation that conditions on both generation model capability and RL iteration to produce calibrated preference scores across the full capability-iteration spectrum. We also introduce HPDv3++, a 212K dual-dimension preference dataset annotated for text-following fidelity and aesthetic quality on frontier generation models. Built upon HPSv3, our two-stage training framework first applies Orthogonal Gradient Descent to absorb new knowledge while preserving existing preference modeling capability, and then performs std-guided semi-supervised adaptive training on unlabeled rollout data spanning diverse model capabilities and RL iterations to learn condition-aware scoring. Experiments show that HPSv3++ achieves state-of-the-art preference prediction accuracy on multiple benchmarks, including a 9.8\% improvement over HPSv3 on HPDv3, while providing stable and effective reward signals for downstream RL fine-tuning across heterogeneous generation models.

%% References
\bibliographystyle{ACM-Reference-Format}
\bibliography{references}

%% Appendix
\clearpage
\appendix
%%%%%%%%%%%%%%%%%%%%%%%%%%%%%%%%%%%%%%%%%%%%%

\section{More RL Fine-tuning Results Across Models and Steps}
\label{sec:appendix_more_rl_results}

\begin{figure*}[t]
    \centering
    \includegraphics[width=0.7\textwidth]{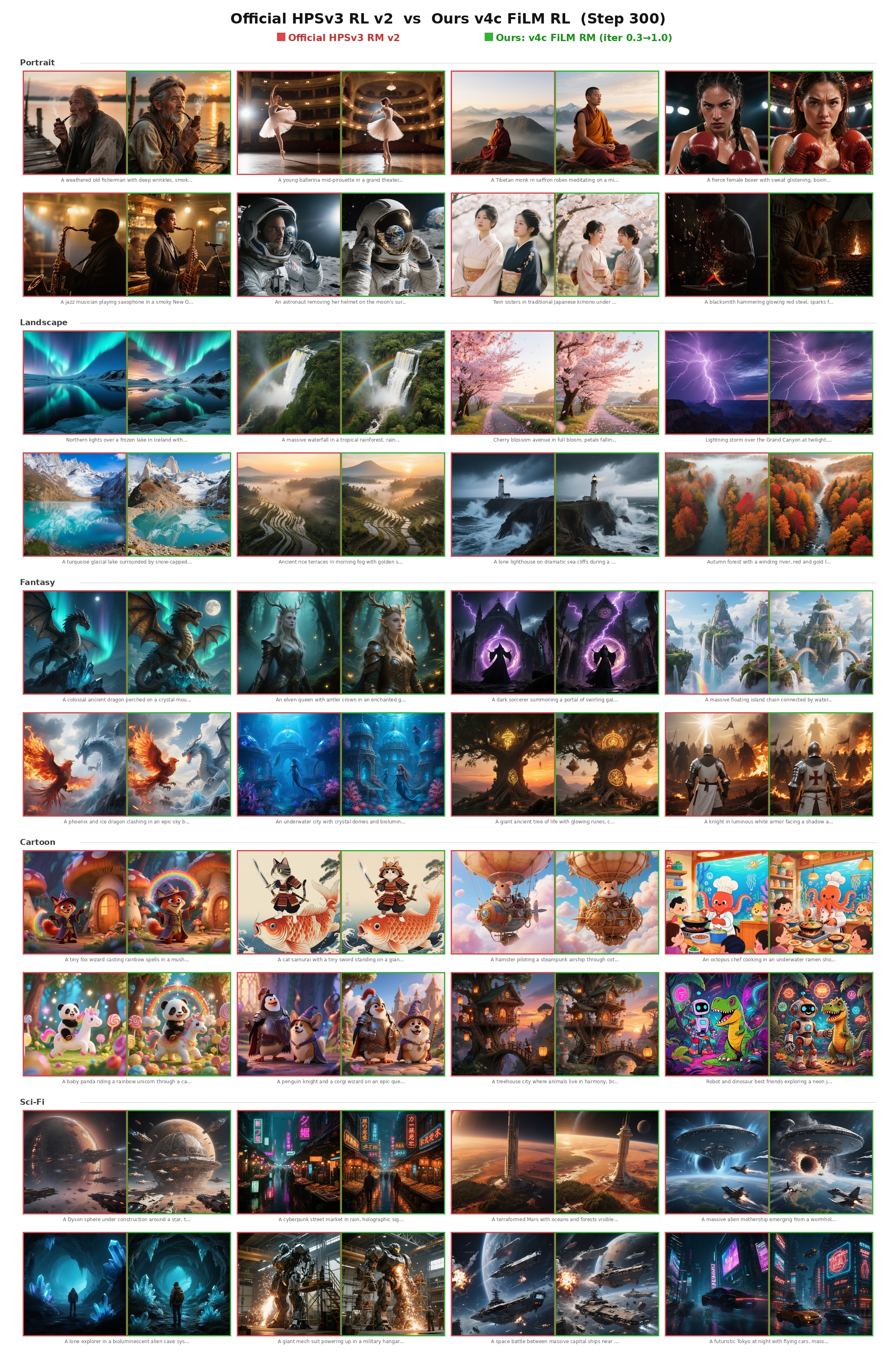}
    \caption{Controlled qualitative comparison under matched RL hyperparameters (Flow-GRPO-based, Qwen-Image, step 300). We compare images generated by models trained with the official HPSv3 reward model (red) and our HPSv3++ reward model (green) across five categories: portrait, landscape, fantasy, cartoon, and sci-fi. HPSv3++ generally yields better prompt fidelity and visual quality across diverse content types.}
    \label{fig:rl_qual_compare}
\end{figure*}

\paragraph{Experimental Setup.}
In this section, we provide additional RL fine-tuning results across different backbones and training steps. We consider three representative text-to-image backbones, namely Qwen-Image, FLUX.1-dev, and SDXL~\cite{qwenimage2025,flux2024,podell2024sdxl}. For all three backbones, we use the PickScore-SFW dataset as the prompt source for RL training, which contains 15,486 prompts. During RL optimization, we sample 8 rollout images per prompt for group-based training, and for evaluation we sample 500 prompts and report quantitative scores from step 0 to step 500. For HPSv3++, the iteration condition is set from 0.3 to 1.0 during RL training. For Qwen-Image, we use a more aggressive RL setting with clip range 0.03 and $\beta=0.001$. For FLUX.1-dev and SDXL, we use a milder setting with clip range 0.003, $\beta=0$, and noise level 0.9, while global std normalization is additionally enabled for SDXL. All comparisons are conducted under matched RL settings, differing only in the reward model. Beyond the quantitative results, we also provide a controlled qualitative comparison in Fig.~\ref{fig:rl_qual_compare}, where we compare Qwen-Image checkpoints at step 300 trained with the official HPSv3 reward model and our HPSv3++ reward model under matched RL hyperparameters. We evaluate the RL-tuned models using four complementary metrics: HPSv2, Aesthetic Score, CLIPScore, and GenEval. HPSv2 reflects human preference consistency, Aesthetic Score evaluates visual aesthetics, CLIPScore measures text-image alignment, and GenEval measures compositional generation ability~\cite{wu2023hpsv2,schuhmann2022aesthetic,radford2021clip,geneval2024}. Table~\ref{tab:rl_all_results} summarizes the full quantitative results.

\begin{table*}[t]
\centering
\caption{Additional RL fine-tuning results across different backbones and training steps. We compare HPSv3++ and HPSv3 as reward models using HPSv2, Aesthetic Score, CLIPScore, and GenEval. Scores are reported on 500 sampled prompts. Higher is better for all metrics. For each backbone, metric, and step, the better result between HPSv3 and HPSv3++ is highlighted in bold, except at step 0.}
\label{tab:rl_all_results}
\scriptsize
\begin{tabular}{l|l|l|cccccc}
\toprule
Backbone & Reward Model & Metric & Step 0 & Step 100 & Step 200 & Step 300 & Step 400 & Step 500 \\
\midrule
\multirow{8}{*}{Qwen-Image}
& \multirow{4}{*}{HPSv3}
& HPSv2            & 0.2883 & 0.2983 & 0.3013 & 0.3042 & 0.2981 & 0.3054 \\
& & Aesthetic Score & 5.8850 & 6.0190 & 5.8150 & 5.8790 & 5.9170 & 5.9910 \\
& & CLIPScore       & 0.2808 & 0.2803 & 0.2817 & 0.2787 & \textbf{0.2765} & \textbf{0.2761} \\
& & GenEval         & 0.9466 & 0.9553 & 0.9616 & 0.9642 & \textbf{0.9680} & \textbf{0.9698} \\
\cline{2-9}
& \multirow{4}{*}{HPSv3++}
& HPSv2            & 0.2883 & \textbf{0.3123} & \textbf{0.3220} & \textbf{0.3345} & \textbf{0.3191} & \textbf{0.3149} \\
& & Aesthetic Score & 5.8850 & \textbf{6.0260} & \textbf{6.0030} & \textbf{6.1750} & \textbf{6.1070} & \textbf{6.0650} \\
& & CLIPScore       & 0.2808 & \textbf{0.2844} & \textbf{0.2819} & \textbf{0.2807} & 0.2707 & 0.2609 \\
& & GenEval         & 0.9466 & \textbf{0.9643} & \textbf{0.9713} & \textbf{0.9759} & \textbf{0.9680} & 0.9677 \\
\midrule
\multirow{8}{*}{FLUX.1-dev}
& \multirow{4}{*}{HPSv3}
& HPSv2            & 0.2769 & 0.2908 & 0.3022 & 0.3135 & 0.3237 & \textbf{0.3293} \\
& & Aesthetic Score & 6.0023 & 6.1020 & 6.1643 & 6.2382 & 6.3158 & \textbf{6.5077} \\
& & CLIPScore       & 0.2705 & \textbf{0.2727} & \textbf{0.2708} & \textbf{0.2706} & \textbf{0.2690} & \textbf{0.2629} \\
& & GenEval         & 0.8698 & \textbf{0.8784} & \textbf{0.8933} & 0.8800 & 0.8699 & 0.8254 \\
\cline{2-9}
& \multirow{4}{*}{HPSv3++}
& HPSv2            & 0.2769 & \textbf{0.2915} & \textbf{0.3064} & \textbf{0.3206} & \textbf{0.3252} & 0.3267 \\
& & Aesthetic Score & 6.0023 & \textbf{6.1111} & \textbf{6.2287} & \textbf{6.3002} & \textbf{6.3197} & 6.4106 \\
& & CLIPScore       & 0.2705 & 0.2705 & 0.2704 & 0.2650 & 0.2610 & 0.2459 \\
& & GenEval         & 0.8698 & 0.8764 & 0.8871 & \textbf{0.9039} & \textbf{0.9080} & \textbf{0.9029} \\
\midrule
\multirow{8}{*}{SDXL}
& \multirow{4}{*}{HPSv3}
& HPSv2            & 0.2883 & 0.2813 & 0.2769 & \textbf{0.2784} & 0.2769 & 0.2748 \\
& & Aesthetic Score & 5.8850 & 6.2110 & 6.1900 & \textbf{6.1990} & \textbf{6.2100} & 6.1800 \\
& & CLIPScore       & 0.2808 & 0.2988 & 0.2967 & 0.2960 & 0.2938 & \textbf{0.2926} \\
& & GenEval         & 0.8687 & 0.8706 & 0.8480 & 0.8362 & 0.8065 & 0.8034 \\
\cline{2-9}
& \multirow{4}{*}{HPSv3++}
& HPSv2            & 0.2883 & \textbf{0.2815} & \textbf{0.2794} & 0.2760 & \textbf{0.2781} & \textbf{0.2810} \\
& & Aesthetic Score & 5.8850 & \textbf{6.2170} & \textbf{6.2070} & 6.1860 & 6.1910 & \textbf{6.2030} \\
& & CLIPScore       & 0.2808 & \textbf{0.2993} & \textbf{0.2976} & \textbf{0.2961} & \textbf{0.2947} & 0.2925 \\
& & GenEval         & 0.8687 & \textbf{0.8731} & \textbf{0.8562} & \textbf{0.8432} & \textbf{0.8438} & \textbf{0.8309} \\
\bottomrule
\end{tabular}
\end{table*}

\paragraph{Analysis.}
Several observations can be drawn from Table~\ref{tab:rl_all_results}. Overall, HPSv3++ maintains strong and stable performance across generation backbones with different capability levels and across different stages of RL optimization, suggesting that capability-aware and iteration-aware reward modeling provides more reliable optimization signals for RL fine-tuning. The advantage is most evident on the high-capability Qwen-Image backbone, where HPSv3++ consistently outperforms HPSv3 on HPSv2 and Aesthetic Score throughout the optimization trajectory, and also achieves stronger GenEval performance during the main RL stages, reaching a higher best score than HPSv3. This indicates that even when the generation backbone already has strong native capability, HPSv3++ can still provide effective reward guidance throughout RL optimization. On the medium-capability FLUX.1-dev backbone, the advantage of HPSv3++ becomes more pronounced in the middle and late stages of RL, especially on GenEval, where it achieves both a stronger trajectory and a higher peak score than HPSv3. On the relatively weaker SDXL backbone, although the margins on HPSv2, Aesthetic Score, and CLIPScore are more modest, HPSv3++ remains consistently better on GenEval across all evaluated checkpoints, demonstrating stronger robustness under a weaker backbone and a more conservative RL regime. The qualitative comparison in Fig.~\ref{fig:rl_qual_compare} further supports these trends. Overall, these results show that HPSv3++ can generalize effectively across different model capability levels and maintain strong performance across different RL iterations, with the clearest and most stable gains appearing on Qwen-Image.

\paragraph{Hyperparameter Discussion.}
From the perspective of RL hyperparameters, Qwen-Image, FLUX.1-dev, and SDXL are all sensitive to the optimization setting. Under more aggressive configurations, both HPSv3 and HPSv3++ become more prone to reward hacking, where numerical metrics may improve without corresponding gains in actual visual quality or human preference alignment. In contrast, under more conservative settings that more closely follow the original Flow-GRPO setup, the numerical differences between the two reward models remain, and our method is typically still slightly better, but the visual differences become less obvious under manual inspection. This suggests that the current hyperparameter regime is more effective at exposing the behavioral differences between reward models during RL, thereby making the advantages of HPSv3++ across different model capabilities and different RL stages more clearly observable.

%%%%%%%%%%%%%%%%%%%%%%%%%%%%%%%%%%%%%%%%%%%%%%%%%

\section{Training and Implementation Details}
\label{sec:appendix_training_details}

This section provides additional implementation details of HPSv3++, focusing on concrete training configurations, module settings, optimization strategies, and training schedules beyond the main paper.

\subsection{Stage~1 Training Details}
\label{sec:appendix_stage1_details}

Stage~1 performs continual learning on HPDv3++ to extend the reward model toward frontier generation model distributions while preserving the original preference modeling capability of HPSv3~\cite{ma2025hpsv3}. We adopt Qwen3-VL-8B-Instruct as the backbone~\cite{bai2025qwen3vltechnicalreport} and use a three-layer RankNet-style reward head on top of the pooled multimodal representation. In our implementation, the pooled hidden feature has dimension 4096, and special-token pooling is used to obtain the representation for reward prediction. The model is initialized from an HPSv3 checkpoint already adapted to the Qwen3-VL backbone.

For training data, we use 191,466 pairwise samples from HPDv3++, including 100,463 aesthetic pairs and 91,003 text-following pairs. As the reference data for Orthogonal Gradient Descent (OGD), we use the HPDv3 training set, which contains 285K pairs. All model parameters are trainable in this stage, including the vision encoder, language model, visual merger, and reward head. At each training step, we sample one batch from HPDv3++ and one batch from HPDv3. The HPDv3++ batch is used to compute the new gradient, while the HPDv3 batch is used to compute the reference gradient for OGD. When the two gradients are compatible, the new gradient is applied directly; otherwise, it is projected onto the orthogonal complement of the reference gradient before the update, following the standard OGD formulation for mitigating interference in continual learning~\cite{lopez2017ogd}. Training is implemented with DeepSpeed ZeRO-2 on 8 NVIDIA H200 GPUs.

\begin{table}[t]
\centering
\caption{Stage~1 model setup.}
\label{tab:appendix_stage1_model}
\small
\begin{tabular}{lp{0.54\linewidth}}
\toprule
Component & Setting \\
\midrule
Backbone & Qwen3-VL-8B-Instruct \\
Hidden size & 4096 \\
Pooling & Special-token pooling \\
Reward head & 3-layer RankNet-style MLP \\
Initialization & HPSv3 checkpoint adapted to Qwen3-VL \\
Trainable modules & Vision encoder, LM, merger, reward head \\
\bottomrule
\end{tabular}
\end{table}

\begin{table}[t]
\centering
\caption{Stage~1 data setup.}
\label{tab:appendix_stage1_data}
\small
\begin{tabular}{lp{0.54\linewidth}}
\toprule
Item & Setting \\
\midrule
Training data & 191,466 pairs from HPDv3++ \\
Aesthetic pairs & 100,463 \\
Text-following pairs & 91,003 \\
Reference data & 285K pairs from the HPDv3 training set \\
OGD batch source & One HPDv3++ batch + one HPDv3 batch per step \\
\bottomrule
\end{tabular}
\end{table}

\begin{table}[t]
\centering
\caption{Stage~1 optimization hyperparameters.}
\label{tab:appendix_stage1_hparams}
\small
\begin{tabular}{lc}
\toprule
Hyperparameter & Value \\
\midrule
Per-device batch size & 2 \\
Gradient accumulation steps & 4 \\
Effective batch size & 64 \\
Reference batch size & 2 per device \\
Epochs & 1 \\
Learning rate & $5\times10^{-7}$ \\
LR schedule & Constant with warmup \\
Warmup ratio & 0.05 \\
Optimizer & AdamW \\
Gradient clipping & 1.0 \\
Precision & bf16 \\
LoRA & Disabled \\
Framework & DeepSpeed ZeRO-2 \\
Hardware & 8 $\times$ NVIDIA H200 \\
Training time & $\sim$2 hours \\
\bottomrule
\end{tabular}
\end{table}

\subsection{Stage~2 Training Details}
\label{sec:appendix_stage2_details}

Stage~2 performs semi-supervised adaptive training with capability- and iteration-conditioned reward modeling. We freeze the vision encoder and language model, and train only the visual merger, reward head, and Capability Encoder. The conditioning module follows a FiLM-hybrid design~\cite{perez2018film}, where the Capability Encoder predicts a capability condition from image features and the normalized RL iteration is injected explicitly into the reward head together with the capability condition. The FiLM condition encoder uses a hidden dimension of 256, and the final FiLM generation layer is zero-initialized so that the conditioned model is initially equivalent to the Stage~1 model.

The Capability Encoder adopts \emph{per-group} capability prediction, i.e., all images within the same rollout group share one inferred capability value. In our final setting, the Capability Encoder is randomly initialized and trained jointly during Stage~2. For labeled supervision, we use the aesthetic subset of HPDv3++, containing 111,650 pairs. For unlabeled rollout data, we use 11,914 multi-model prompts, each rendered by three capability tiers (Stable Diffusion~1.5, SDXL, and Qwen-Image), yielding 35,742 groups in total. We additionally use 3,000 multi-iteration prompts, evenly split between SDXL and FLUX.1-dev, and sample six RL checkpoints from step 0 to step 1000 using rollouts generated with the Stage~1 OGD model as the reward signal, yielding 18,000 groups. Together, these two sources provide 53,742 unlabeled training groups. The source models involved in Stage~2 are Stable Diffusion~1.5, SDXL, FLUX.1-dev, and Qwen-Image~\cite{podell2024sdxl,flux2024,qwenimage2025}. Specifically, the multi-model groups are constructed from Stable Diffusion~1.5, SDXL, and Qwen-Image, while the multi-iteration groups are sampled from RL checkpoints of SDXL and FLUX.1-dev along the Flow-GRPO trajectory~\cite{flowgrpo2025}. Stage~2 is initialized from the OGD checkpoint obtained in Stage~1.

Each training step combines three data streams from separate dataloaders: an unlabeled group batch, a cross-model ranking batch, and a labeled preference batch. The total Stage~2 objective is
\begin{equation}
\begin{aligned}
\mathcal{L}_{\text{total}} ={}&
\lambda_{\text{sup}}\mathcal{L}_{\text{sup}}
+
\lambda_{\text{rank}}\mathcal{L}_{\text{rank}}
+
\lambda_{\text{std}}\mathcal{L}_{\text{std}}\\
&+
\lambda_{\text{adapt}}\mathcal{L}_{\text{adapt}}
+
\eta_{\text{constraint}}\mathcal{L}_{\text{constraint}}
+
\eta_{\text{pred}}\mathcal{L}_{\text{pred}}
+
\eta_{2}\mathcal{L}_{2},
\end{aligned}
\end{equation}
where $\mathcal{L}_{\text{sup}}$ is the supervised uncertainty-aware preference loss on labeled HPDv3++ pairs, $\mathcal{L}_{\text{rank}}$ is a Bradley--Terry ranking loss computed on cross-model ranking pairs, $\mathcal{L}_{\text{std}}$ is the std-promotion term, $\mathcal{L}_{\text{adapt}}$ is the adaptive ratio-ordering term, $\mathcal{L}_{\text{pred}}$ is a smooth-$L_1$ supervision term on the predicted condition values, and $\mathcal{L}_{2}$ is a weak reward-magnitude regularizer. Since $\mathcal{L}_{\text{std}}$ and $\mathcal{L}_{\text{adapt}}$ are already introduced in the main text, we only detail the ratio-constraint term here.

Let $\rho_g=\sigma_g/\sigma_g^{\text{base}}$ denote the std ratio of group $g$, where $\sigma_g$ is the group score std of the current model and $\sigma_g^{\text{base}}$ is the corresponding std produced by the Stage~1 model. We define
\begin{equation}
\mathcal{L}_{\text{constraint}}
=
\mathcal{L}_{\text{bound}} + 5\,\mathcal{L}_{\text{target}},
\end{equation}
with
\begin{equation}
\mathcal{L}_{\text{bound}}
=
\frac{1}{G}\sum_{g=1}^{G}
\Bigl[
\operatorname{ReLU}(\ell_g-\rho_g)
+
\operatorname{ReLU}(\rho_g-u_g)
\Bigr],
\end{equation}
\begin{equation}
\mathcal{L}_{\text{target}}
=
\frac{1}{G}\sum_{g=1}^{G}
\operatorname{ReLU}\!\bigl(|\rho_g-\tau_g|-\delta_g\bigr),
\end{equation}
where $\ell_g$, $u_g$, $\tau_g$, and $\delta_g$ are iteration-dependent lower bound, upper bound, target, and margin terms. In our final setting, these are
\begin{equation}
\ell_g = 0.85 + 0.25\,t_g,
\qquad
u_g = 1.15 + 0.45\,t_g,
\end{equation}
\begin{equation}
\tau_g = 1.0 + 0.4\,t_g,
\qquad
\delta_g = 0.1 + 0.05\,t_g,
\end{equation}
where $t_g$ is the normalized RL iteration value of group $g$. This term prevents the std ratio from collapsing or growing without bound, while still allowing later RL stages to exhibit larger relative std gains. In the final model, the ranking term is computed only over cross-model pairs, and the Capability Encoder is trained jointly from random initialization. Stage~2 is implemented with DeepSpeed ZeRO-2 on 8 NVIDIA H200 GPUs.

\begin{table}[t]
\centering
\caption{Stage~2 model setup.}
\label{tab:appendix_stage2_model}
\small
\begin{tabular}{lp{0.54\linewidth}}
\toprule
Component & Setting \\
\midrule
Initialization & Stage~1 OGD checkpoint \\
Model type & FiLM hybrid \\
Condition hidden dim & 256 \\
Capability prediction & Per-group \\
Trainable modules & Merger, reward head, Capability Encoder \\
Frozen modules & Vision encoder, language model \\
Capability Encoder setting & Random init + trainable \\
Ranking pair mode & Cross-model only \\
Predicted-cond supervision & 1.5 (smooth-$L_1$) \\
\bottomrule
\end{tabular}
\end{table}

\begin{table}[t]
\centering
\caption{Stage~2 data setup.}
\label{tab:appendix_stage2_data}
\small
\begin{tabular}{lp{0.54\linewidth}}
\toprule
Item & Setting \\
\midrule
Labeled data & 111,650 HPDv3++ aesthetic pairs \\
Unlabeled prompts & $\sim$12K multi-model prompts + $\sim$3K multi-iteration prompts \\
Expanded groups & $\sim$54K groups in total \\
Source models & SD~1.5, SDXL, FLUX.1-dev, Qwen-Image \\
Batch organization & Separate labeled, ranking, and unlabeled dataloaders \\
\bottomrule
\end{tabular}
\end{table}

\begin{table}[t]
\centering
\caption{Stage~2 optimization hyperparameters.}
\label{tab:appendix_stage2_hparams}
\small
\begin{tabular}{lc}
\toprule
Hyperparameter & Value \\
\midrule
Per-device batch size & 4 \\
Gradient accumulation steps & 1 \\
Effective batch size & 32 \\
Epochs & 2 \\
Learning rate & $1\times10^{-5}$ \\
Merger learning rate & $5\times10^{-6}$ \\
LR schedule & Cosine \\
Warmup ratio & 0.05 \\
Gradient clipping & 1.0 \\
Precision & bf16 \\
$\lambda_{\text{sup}}$ & 5.0 \\
$\lambda_{\text{rank}}$ & 2.0 \\
$\lambda_{\text{std}}$ & 1.0 \\
$\lambda_{\text{adapt}}$ & 2.0 \\
$\eta_{\text{constraint}}$ & 3.0 \\
$\eta_{\text{pred}}$ & 1.5 \\
$\eta_{2}$ & 0.03 \\
Constraint warmup & 300 steps \\
Std/adapt/pred warmup & 500 steps \\
Framework & DeepSpeed ZeRO-2 \\
Hardware & 8 $\times$ NVIDIA H200 \\
\bottomrule
\end{tabular}
\end{table}

\begin{table}[t]
\centering
\caption{Capability Encoder ablation (\%).}
\label{tab:appendix_capencoder}
\small
\begin{tabular}{lccc}
\toprule
Strategy & HPDv3 & Aesthetic & T-Fol. \\
\midrule
Pretrained + frozen & 77.8 & 75.1 & 80.9 \\
Pretrained + trainable & 81.8 & 77.1 & 85.3 \\
Random init + trainable & \textbf{86.7} & \textbf{79.1} & \textbf{88.1} \\
\bottomrule
\end{tabular}
\end{table}
%%%%%%%%%%%%%%%%%%%%%%%%%%%%%%%%%%%%%%%%%%

\section{HPDv3++ Dataset Construction Details}
\label{sec:appendix_dataset_details}

HPDv3++ is constructed to broaden preference supervision toward frontier text-to-image generation distributions while providing finer-grained annotations than prior reward-model datasets~\cite{wu2023hps,ma2025hpsv3,xu2024imagereward,kirstain2023pickapic}. Compared with HPDv3, the key difference is that HPDv3++ is built on prompts paired with images generated by a recent high-capability model, and is annotated separately for text-following fidelity and aesthetic quality.

\begin{figure}[t]
    \centering
    \includegraphics[width=0.95\linewidth]{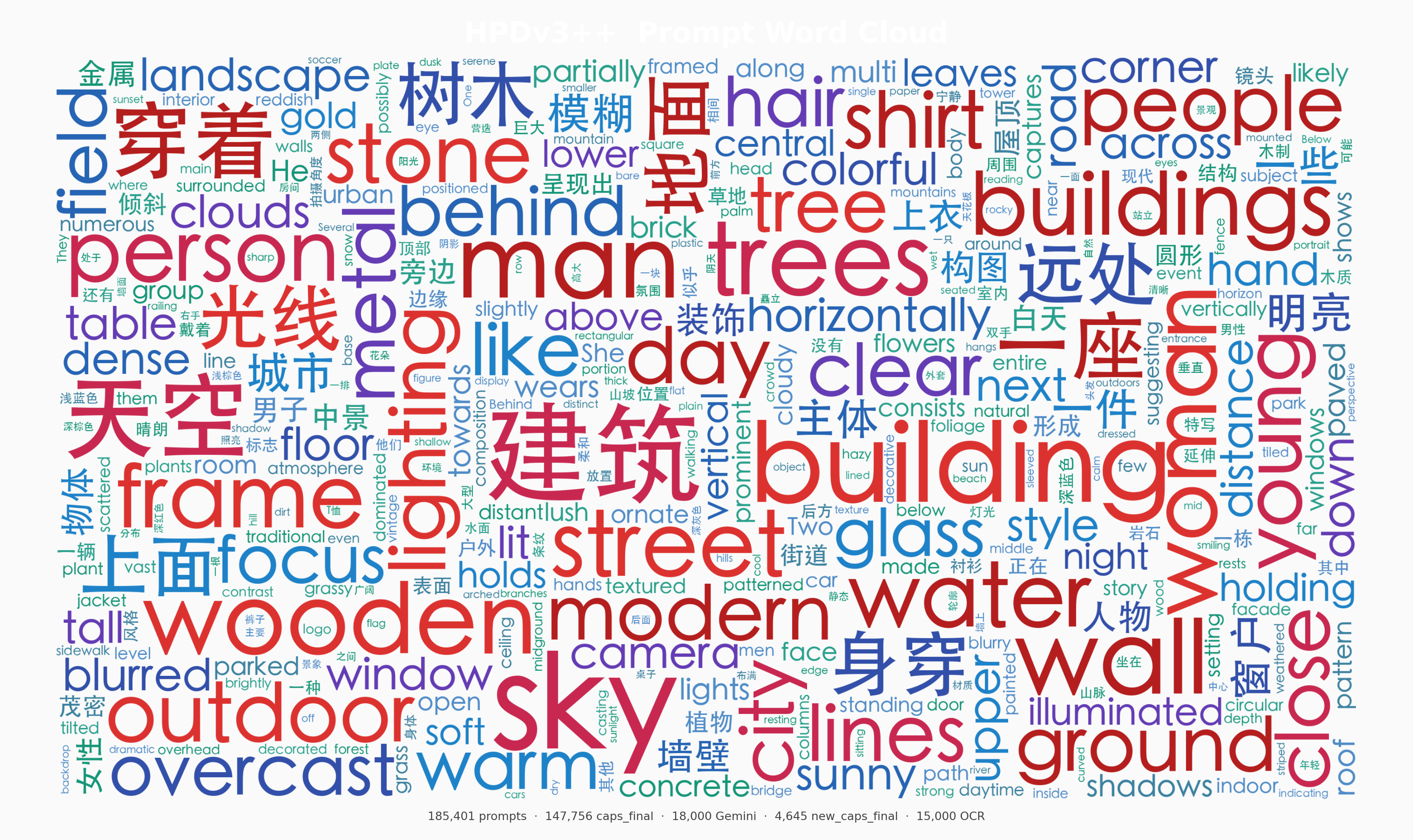}
    \caption{Word cloud of the collected prompts in HPDv3++, illustrating the broad semantic coverage of the dataset and the increased presence of person-centric, activity-related, and text-rich concepts.}
    \label{fig:prompt_cloud}
\end{figure}

\begin{figure*}[t]
    \centering
    \includegraphics[width=0.98\textwidth]{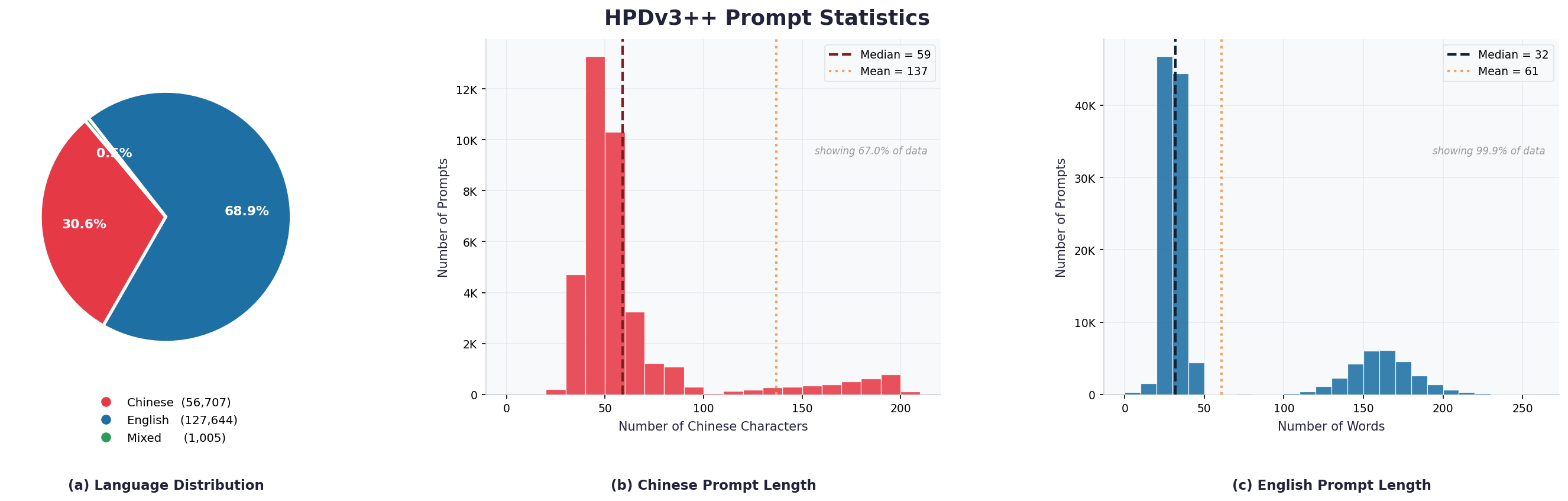}
    \caption{Prompt statistics of HPDv3++. Left: language distribution of the collected prompts. Middle: length distribution of Chinese prompts measured by the number of Chinese characters. Right: length distribution of English prompts measured by the number of words. The dataset contains a substantial proportion of English prompts while also preserving a large Chinese subset, and exhibits a long-tailed prompt-length distribution that covers both short prompts and highly compositional descriptions.}
    \label{fig:prompt_stats}
\end{figure*}

\subsection{Prompt Preparation and Image Generation}
\label{sec:appendix_prompt_generation}

We collect prompts from multiple sources and organize them into three broad categories. The first category consists of natural image-caption corpora derived from real images, including both the main caption source and its later supplements. These prompts are predominantly in Chinese and cover a wide range of scenarios such as landscapes, architecture, indoor scenes, objects, people, and daily activities. The second category consists of Gemini-generated captions, which are also based on real-image descriptions but place stronger emphasis on people, sports, and activity-centric scenes. The third category consists of OCR-oriented prompts that explicitly describe images containing overlaid text or text-rendering requirements, and are introduced to improve coverage of text rendering ability.

As illustrated in Fig.~4, these prompt sources are intentionally designed to complement each other. In particular, we substantially increase the proportion of activity- and sports-related prompts, which were relatively underrepresented in earlier data construction pipelines. This supplementation better reflects the types of content that are increasingly emphasized in newer generation models, including human activities, dynamic body poses, interaction-heavy scenes, and text-rich visual compositions. We further visualize the overall prompt semantics using a word cloud in Fig.~\ref{fig:prompt_cloud}, which highlights the broad semantic coverage of the collected prompts and also reveals the increased presence of person-centric, activity-related, and text-rich concepts in HPDv3++.

As further shown in Fig.~\ref{fig:prompt_stats}, the final prompt collection contains 185,401 prompts in total, including 127,644 English prompts (68.9\%), 56,707 Chinese prompts (30.6\%), and 1,005 mixed-language prompts (0.5\%). The middle and right panels of Fig.~\ref{fig:prompt_stats} also show that the prompt-length distribution is highly diverse. For Chinese prompts, the median length is 59 characters and the mean length is 137 characters, indicating a substantial long-tail of detailed descriptions. For English prompts, the median length is 32 words and the mean length is 61 words, again reflecting a mixture of short prompts and highly compositional instructions. These statistics further confirm that HPDv3++ covers not only a broad semantic range, but also diverse prompt styles and complexity levels.

After collection, we perform semantic deduplication with Qwen3-Embedding-8B. Prompts are processed sequentially without shuffling, and a new prompt is removed if its cosine similarity to any retained prompt is at least 0.75. We then categorize the remaining prompts into four macro groups, namely Nature, Design, People, and Synthetic, together with their finer subcategories. We further assign language tags and length buckets, and keep the higher-quality subset within each category to improve both diversity and prompt quality. After this filtering process, we obtain approximately 185K prompts spanning more than 20 scenarios.

For each retained prompt, we generate six candidate images with Qwen-Image, yielding over 1.1M candidate images in total. We use Qwen-Image because it represents a recent high-capability generator that is absent from the training distribution of HPDv3, thereby extending preference supervision toward frontier generation models~\cite{qwenimage2025}. In image generation, we use a classifier-free guidance scale of 4 and set \texttt{num\_inference\_steps} to 25.

\subsection{Annotation Protocol}
\label{sec:appendix_annotation_protocol}

To ensure consistent and fine-grained human supervision, we annotate each prompt along two complementary dimensions. The first is \emph{text-following fidelity}, which evaluates how faithfully an image reflects the prompt semantics, including object presence, attribute accuracy, spatial relationships, action or scene consistency, viewpoint, text rendering, and layout constraints. The second is \emph{aesthetic quality}, which evaluates visual appeal and perceptual quality, including image clarity, composition, lighting, color harmony, anatomical plausibility, physical consistency, texture realism, and the absence of obvious artifacts or strong AI-generated visual traces.

For each prompt, annotators are presented with the six generated images and are asked to identify, separately for each dimension, the \emph{best} image and the \emph{worst} image. If the candidates are judged to be indistinguishable for a given dimension, that group is discarded for that dimension. Otherwise, a pairwise preference sample is constructed from the selected best--worst pair.

For text-following fidelity, annotators are instructed to focus primarily on whether the prompt requirements are satisfied, including object counts, actions, spatial relations, left--right orientation, scene layout, gaze direction, and text content when applicable. For text-rich prompts, additional attention is paid to rendering details such as quotation marks, punctuation, line arrangement, and short-keyword fidelity. Prompts that are incomplete, self-contradictory, or semantically invalid are marked as tie cases and excluded.

For aesthetic quality, annotators are instructed to focus on realism, structural correctness, and perceptual fidelity. Common failure cases include broken body structure, implausible motion, contact inconsistencies, repeated background patterns, edge artifacts, unreadable text, fake logos, over-smoothed skin, and visually noisy yet semantically meaningless details. In this dimension, severe structural errors take priority over weaker AI-like artifacts, and images with clear structural collapse cannot be selected as the best sample. We also place additional emphasis on activity- and sports-related images, where body pose, motion plausibility, and interaction consistency are especially important.

\subsection{Data Cleaning, Quality Verification, and Final Statistics}
\label{sec:appendix_dataset_cleaning}

Following the above protocol, human annotators provide pairwise supervision for both aesthetic quality and text-following fidelity. To improve reliability, the annotations are further reviewed by annotation group leaders and additionally spot-checked by technical staff. Before automatic filtering, the collected annotations yield approximately 145K text-following pairs and 164K aesthetic pairs.

After annotation, we further clean the data in two ways. First, we remove noisy prompts that contain meaningless random character strings or explicit text-layout instructions, while preserving naturally occurring scene text such as signs or road text. Second, we perform pair deduplication to remove repeated samples, including duplicate prompt--pair combinations, different image pairs under the same prompt, and identical image pairs associated with different prompts.

We further apply a conservative dual-judge verification procedure for automatic quality control. A Qwen3-VL-32B-Instruct model serves as a VLM judge, and the original HPSv3 model serves as an RM judge~\cite{bai2025qwen3vltechnicalreport,ma2025hpsv3}. The VLM judge is applied separately to the aesthetic and text-following dimensions, and the input order of image pairs is randomized to reduce position bias. A sample is discarded only when \emph{both} the VLM judge and the HPSv3 judge disagree with the human annotation. This conservative strategy preserves the majority of human-labeled samples while filtering out obvious annotation errors.

After prompt cleaning, pair deduplication, and dual-judge verification, the final dataset contains approximately 95K text-following pairs and 117K aesthetic pairs, for a total of approximately 212K pairs. Overall, HPDv3++ is a dual-dimension preference dataset built on frontier-model generations, with expanded coverage of high-capability image distributions and stronger emphasis on activity, sports, and text-rich scenarios that are increasingly important in modern text-to-image generation.

%%%%%%%%%%%%%%%%%%%%%%%%%%%%%%%%%%%%%%%%%%%%%%%%%%

\section{Limitations}
\label{sec:appendix_limitations}

Despite its improved coverage over model capability and RL iteration, HPSv3++ still has several limitations that may affect its generalizability in broader settings.

\subsection{Dataset Scope and Annotation Noise}

HPDv3++ is constructed from prompts paired with images generated by a single frontier model, namely Qwen-Image~\cite{qwenimage2025}. Although this design helps us extend preference supervision toward stronger generation distributions, it also means that the dataset may not fully cover the diversity of outputs produced by all modern text-to-image systems. In particular, while we intentionally strengthen the coverage of activity-, sports-, and text-rich prompts to address weaknesses in earlier data construction and better match the content emphasized by newer generators, the resulting distribution is still centered on general-purpose image generation rather than specialized domains such as medicine, scientific visualization, or other expert-oriented applications.

Our annotation pipeline explicitly separates text-following fidelity and aesthetic quality, which already provides finer supervision than single-score preference datasets~\cite{wu2023hps,xu2024imagereward}. However, human preference in text-to-image generation can also depend on factors that are not directly modeled here, such as style faithfulness, cultural context, safety-related judgments, or domain-specific correctness. Moreover, although we adopt structured annotation protocols together with conservative dual-judge filtering, preference annotation remains inherently subjective, especially in ambiguous cases or when text-following fidelity and aesthetic quality conflict. Such cases may still introduce residual noise into the dataset.

\subsection{Capability-Iteration Coverage and RL Robustness}

Although Stage~2 introduces capability- and iteration-aware conditioning, the rollout data used for semi-supervised adaptation still covers only a limited portion of the full capability-iteration space. In practice, it may not fully represent unseen generators, substantially longer RL horizons, or future generation paradigms. As a result, the learned conditioning mechanism should be viewed as an important step toward broader reward calibration, rather than a complete solution for all possible generation distributions.

In addition, while HPSv3++ provides more stable and effective reward signals than HPSv3 in our experiments, it does not completely eliminate residual reward exploitation or overoptimization in challenging RL settings. In particular, downstream generators may still exploit blind spots of the reward model in cases involving subtle structural errors, highly repetitive patterns, or distribution shifts beyond the current rollout construction.

\end{document}